%% file: main_arxiv.tex
\documentclass[11pt]{article}

\usepackage[preprint]{acl}
\usepackage{times}

\usepackage{latexsym}
\usepackage[T1]{fontenc}
\usepackage[utf8]{inputenc}
\usepackage{microtype}
\usepackage{graphicx}
\usepackage{booktabs}
\usepackage{tabularx}
\usepackage{placeins}
\usepackage{amsmath}
\usepackage{amssymb}

\title{The Limits of Speculation: Bounding Speculative Decoding in Mixture-of-Experts}
\author{Aidar Amankulov \And Denis Mamatin}

\begin{document}
\maketitle

\begin{abstract}
Speculative decoding in Mixture-of-Experts (MoE) models faces the problem of unstable verification cost caused by input-dependent expert loading. To study the physics of this process, we formulate speculation-budget selection as an offline Stochastic Shortest Path (SSP) problem over reference sequences and build a diagnostic Oracle that uses counterfactual simulation to account for MoE verification cost. A detailed analysis of the Oracle's decisions on the Qwen3-Coder and EAGLE-3 pairing, in the space of marginal deltas (Delta Space), shows that rejected candidates form a strict linear boundary. This result demonstrates that a complex global optimization is locally governed by a necessary condition balancing marginal cost against expected progress ($\frac{\Delta \mathbb{E}[Cost]}{\Delta \mathbb{E}[a]}$), providing a rigorous mathematical reference point for designing future adaptive online heuristics.
\end{abstract}

\section{Introduction}

Autoregressive decoding remains sequential, and at small batch sizes its latency is often dominated by reading weights and the KV cache from memory \citep{shazeer2019fast,MLSYS2023_c4be71ab}.
Speculative decoding (SD) reduces the number of sequential calls to the target model: a cheap draft model proposes several tokens, which are verified by the target model in a single parallel pass \citep{pmlr-v202-leviathan23a,chen2023acceleratinglargelanguagemodel}.
For dense models this pass partially amortizes the weight read, so generation efficiency is usually tied directly to draft-model quality and the number of accepted tokens.

For Mixture-of-Experts (MoE) architectures, this classical explanation is insufficient.
Because the router selects experts separately for each token \citep{shazeer2017outrageously,fedus2022switch,lepikhin2020gshard}, positions within a single speculative pass can access entirely different sets of weights \citep{huang2025moesd,saxena2025utility}.
A long speculative chain increases the expected number of accepted tokens, but it simultaneously widens the union of activated experts for the tail of positions that may end up being rejected.
Consequently, a high token-acceptance probability alone no longer guarantees a speedup: in MoE models one must carefully balance the chance of a correct guess against the input-dependent cost of verification.

Existing approaches try to address this problem by developing various online heuristics (e.g., confidence-based cutoffs or dynamic budgets). However, because of the high local variance in routing cost, such heuristics often operate blindly, without a rigorous analytical reference point.
Rather than proposing yet another empirical online policy, in this work we take a step back and ask: \textit{what is the absolute theoretical limit on the speedup of speculative decoding in MoE, and what fundamental rules does the ideal strategy obey?}

To answer this question, we formulate the search for the optimal speculation budget as a Stochastic Shortest Path (SSP) problem over a known reference (ground-truth) sequence. We build a hybrid offline Oracle that lets us isolate algorithmic losses from systems-level ones and study the local ``physics'' of decision-making.

\paragraph{Our Contributions.} Our contributions are as follows:
\begin{itemize}
    \item We develop the \textbf{Sequence-Conditioned Offline Oracle} methodology. Using counterfactual simulation of ``junk tails'', the Oracle allows an honest estimate of the cost of erroneous tokens in an MoE model without distorting the reference generation path.
    \item We empirically demonstrate the \textbf{limits of speculation in MoE}. Unlike dense models, increasing the available budget ($L_{max}$) in MoE leads to a strong diminishing-returns effect: the ideal algorithm forcibly truncates budgets to avoid routing penalties for low-probability tokens.
    \item We introduce \textbf{Delta Space Analysis} --- an examination of the Oracle's space of marginal decisions. We show that the complex global search over the Bellman equation reduces in practice to maintaining a simple linear balance ($\frac{\Delta \mathbb{E}[Cost]}{\Delta \mathbb{E}[a]}$). This demonstrates that lightweight online heuristics based on early exit, triggered once a constant cost-slope barrier is crossed, are feasible.
\end{itemize}

\section{Background \& Related Work}
\label{sec:background-related-work}

\subsection{Decode Cost as the Baseline for Speculative Decoding}

Speculative decoding targets the \texttt{decode} phase, in which autoregressive generation sequentially appends tokens after the \texttt{prefill} phase has built the KV cache \citep{NIPS2017_3f5ee243,MLSYS2023_c4be71ab}.
At small batch sizes, each decoding step can be dominated by reading model weights and KV-cache state from HBM, so latency is not captured by \texttt{FLOPs} alone \citep{shazeer2019fast,MLSYS2023_c4be71ab}.

\subsection{MoE Inference}

MoE models increase overall capacity through conditional computation: for each token, only a small subset of experts, chosen by the router, is activated \citep{shazeer2017outrageously,fedus2022switch,lepikhin2020gshard}.
Because routing is performed independently for each token, positions verified in a single target-model pass may access different expert identifiers, which can increase the volume of weight access and overhead on the target-model side \citep{huang2025moesd,saxena2025utility}.
The execution path of the target MoE model remains unchanged; what is measured is the relationship between draft length and the layer-averaged number of unique experts, on one hand, and verification cost, on the other.

\subsection{Speculative Decoding}

Speculative decoding accelerates autoregressive generation via a ``draft-then-verify'' scheme \citep{pmlr-v202-leviathan23a,chen2023acceleratinglargelanguagemodel}.
In each round, the draft model proposes $K$ draft tokens, and the target model verifies them in a single verification pass.
Tokens are then accepted sequentially up to the first mismatch; under sampling, rejection sampling is used to preserve the target model's distribution.

EAGLE is treated here as an example of a strong contemporary draft-side approach \citep{li2024eagle,li2024eagle2,li2025eagle3}.
The key question is a different one: can the target MoE model's verification pass become expensive enough that the expected speedup vanishes even with a sufficiently good draft model?

\subsection{Related Work}

Classical speculative decoding verifies draft-model continuations with the target model and preserves the target model's distribution via rejection sampling \citep{pmlr-v202-leviathan23a,chen2023acceleratinglargelanguagemodel}.
This line of work has produced stronger draft-side mechanisms, including feature-level drafting, dynamic candidate trees, multiple output heads, architectural changes to the draft decoder, and specialized training objectives \citep{li2024eagle,li2024eagle2,li2025eagle3,cai2024medusa,wang2026prism,zhong2025beagle,samarin2026lklosses}.
Such methods mainly improve candidate quality, draft-model overhead, or tree construction prior to verification by the target model.
The present study instead isolates the configuration-dependent cost of the target MoE model's verification pass itself.

Methods with adaptive lookahead depth change the length or structure of speculation at runtime.
DISCO uses a draft-model confidence signal, GammaTune adapts the budget from the acceptance history of previous rounds, SVIP ties stopping to draft entropy, TETRIS optimizes token selection under batching, and AdaEAGLE explicitly predicts an adaptive structure \citep{pmlr-v262-mamou24a,gautam2025gammatune,zhang2024draftmodelknows,wu2025tetris,zhang2024adaeagle}.
These works motivate dynamic control; DISCO and GammaTune, however, do not directly model the MoE-specific signal of expert-union size in the target model's pass.
Consequently, adaptive or oracle-based length selection by itself is not a contribution of the present study.

Work on speculative decoding for MoE explicitly accounts for the cost of target verification.
MoESD introduces target-model efficiency, and Cascade weighs expected progress against verification cost to avoid unfavorable speculation \citep{huang2025moesd,saxena2025utility}.
EVICT and MoE-Spec directly exploit the growth of the activated-expert union: EVICT prunes the candidate tree, and MoE-Spec caps the per-layer expert budget \citep{pan2026evict,mcdanel2026moespec}.
EcoSpec incorporates the predicted marginal cost of expert activation into draft-token selection \citep{xie2026lessexperts}, while Sparse Verification sparsifies the computation of verification itself \citep{wang2025sparseverification}.
These approaches modify the verification policy or the target model's computation; the link between the number of unique experts and cost is therefore not claimed here as novel.

A related line of systems research treats expert loading and data movement as the binding constraint.
SpecMoE, SP-MoE, and SpecMoEOff use self-speculation, weight prefetching, or offloading to reduce the memory burden of MoE inference, while MoE-SpeQ uses quantized drafting for accurate prediction of expert routing and its prefetching, hiding I/O latency \citep{bang2026specmoe,chen2025spmoe,wang2025specmoeoff,wang2025moespeq}.
These systems change the expert-loading regime or the model's execution environment.
SPEED-Bench further shows that results depend on workload, batching, and environment \citep{abramovich2026speedbench}.
By the same logic, the synthetically resampled counterfactual tail in the present work could in principle route differently than the real continuation of the same position.

The contribution of this paper is limited to a configuration-dependent cost calibration of an unmodified target MoE model and to a sequence-conditioned SSP analysis.
The paper does not propose an online policy and does not claim superiority over EVICT, MoE-Spec, EcoSpec, or other adaptive methods.

\section{Problem Formulation \& The Cost Model}
\label{sec:problem-cost-model}

\subsection{Useful Progress and Target Model Time}

Following MoESD \citep{huang2025moesd}, we separate three quantities: mean useful progress, draft cost, and target-model verification time; $B=1$ in all experiments.
For a measured run of $R$ rounds, $G_r$ denotes the number of useful output tokens in round $r$; in the sequence-conditioned setting below, $a$ denotes the number of consecutive matches with a fixed reference and is not a count of rejection-sampling acceptances.
Let $T_T^{(B)}(K)$ denote the target model's verification time for $K$ draft-token positions plus one extra target-model position, and let $T_T^{(B)}(0)$ denote the same pass with no draft tokens.
The normalized target-model efficiency from MoESD isolates the ratio of these pass times:
\begin{equation}
\eta_B(K)=\frac{T_T^{(B)}(0)}{T_T^{(B)}(K)}.
\label{eq:target-efficiency}
\end{equation}
For a fixed batch size and execution-environment configuration, we hereafter allow the shorthand $T_T(K)$ and $\eta(K)$; in the present analysis, $\eta(K)$ serves as a descriptive normalization and does not enter directly into the offline policy's objective.

Uppercase $K$ henceforth denotes the measured calibration budget, while lowercase $k$ denotes the offline policy's action (Section~\ref{sec:offline-oracle}).

Here MoE changes the problem setting: unlike dense models, where $T_T(K)$ depends weakly on $K$ from a memory-traffic standpoint, in MoE each verified token selects its own subset of experts, so the verification pass can increase the layer-averaged number of unique experts $\bar{U}_r$ relative to a single decoding step \citep{huang2025moesd,saxena2025utility}.
This dependence among $T_T(K)$, $\eta(K)$, and $\bar U_r$ is unpacked by the verification cost model below.

\subsection{Speedup and Degradation with Increasing $K$}
\label{sec:speedup-degradation}

Following MoESD \citep{huang2025moesd}, the realized speedup can be written as the ratio of useful progress to the round's total cost. Decomposing the round cost into draft cost and target-model verification time, and dropping the small term accounting for rejection sampling and post-processing, we obtain the approximation
\begin{equation}
\begin{gathered}
\mathrm{SpeedUp}
\approx
\frac{\mathbb{E}[G_r]}
{K\dfrac{T_D}{T_T(0)}+\dfrac{1}{\eta(K)}},\\[4pt]
\mathbb{E}[G_r]=\frac{1}{R}\sum_{r=1}^{R}G_r,
\end{gathered}
\label{eq:speedup-approx}
\end{equation}
where $G_r$ is the number of useful output tokens in round $r$, $\eta(K)$ is defined by Eq.~\eqref{eq:target-efficiency}, and $T_D$ is the cost of a single draft position; the calibrated form of $T_D$ is given in Section~\ref{sec:cost-model-validation}.

As the budget $K$ grows, the numerator can increase through larger useful progress $\mathbb E[G_r]$, but the denominator also grows: the term $KT_D/T_T(0)$ grows linearly in $K$, and $1/\eta(K)$ grows if target efficiency $\eta(K)$ declines. For dense models, the decline of $\eta(K)$ with increasing $K$ is typically weaker; for MoE targets, it is observably linked to growth in the layer-averaged number of unique experts $\bar U_r$ in the verification microbatch, which the verification cost model below unpacks (Section~\ref{sec:cost-model}). Thus, the benefit of increasing $K$ is not guaranteed in advance: it depends on whether the gain in $\mathbb E[G_r]$ offsets the growth of the denominator.

\subsection{Verification Cost Model}
\label{sec:cost-model}

To describe the decline in target-model efficiency as speculation length grows, we account for the hardware constraints of MoE inference.
In dense models, the cost of a verification step is often dominated mainly by the number of tokens verified in parallel.
In MoE models, a substantial component of the forward-pass latency $T_T^{(B)}(K)$ can be memory traffic tied to expert weights.

Each token in the speculative chain is routed to its own token-specific subset of experts.
If the corresponding weights are not reused from fast on-chip memory, reading them from global memory can increase latency.
The layer-averaged number of unique experts across all positions of the pass is therefore treated as an observed cost signal, not as a causally identified volume of traffic.
Within a round, $\bar U_r$ is computed over the same verification positions as $T_T^{(B)}(K)$: unique experts are counted separately in each MoE layer, and the result is then averaged across layers.
In compact form, this intuition corresponds to the dependence
\begin{equation}
T_{T,r}\approx\alpha\bar U_r+\beta,
\label{eq:target-cost-intuition}
\end{equation}
but the actual predictive model below additionally accounts for context and the categorical form of the microbatch.

Let $m_r=K_r+1$ denote the number of positions computed by the target model in round $r$: $K_r$ draft positions plus one extra target-model position.
The batch size remains $B=1$ throughout.
Measured latency depends on three observed channels: the expert union $\bar U_r$, the context length $L_{\mathrm{ctx},r}$, and the shape of the verification microbatch $m_r$.
The first channel is consistent with reading expert weights from HBM, the second captures KV-cache and attention cost, and the third accounts for block size and CUDA-graph mode.

For the configuration under study, we use the diagnostic regression model
\begin{equation}
T_{T,r}
=
\beta_0(m_r)
+
\alpha(m_r)\bar U_r
+
cL_{\mathrm{ctx},r}
+
\varepsilon_r,
\label{eq:target-cost}
\end{equation}
where $\beta_0(m)$ and $\alpha(m)$ are free categorical coefficients for each $m=K+1$, $c$ is the shared context coefficient, and $\varepsilon_r$ is the residual.
The categorical form describes the observed plateaus of $\alpha(m)$ and the non-monotonicity between $m=4$ and $m=5$ better than a shared linear dependence.
The breakpoint was chosen after inspecting the full sweep, so it is a descriptive part of this calibration rather than an independently confirmed boundary of a CUDA-graph mode.

Formally, let $\mathcal I_r$ contain all $m_r$ positions of the verification pass, including the extra target-model position, and let $\mathcal U_{\ell,r}$ be the union of expert identifiers selected for these positions in MoE layer $\ell$.
The routing signal we use is
\begin{equation}
\bar{U}_r
=
\frac{1}{|\mathcal{L}_{\mathrm{MoE}}|}
\sum_{\ell \in \mathcal{L}_{\mathrm{MoE}}}
|\mathcal{U}_{\ell,r}|.
\label{eq:layer-averaged-experts}
\end{equation}
Thus, $\bar U_r$ has the same full-microbatch, per-round granularity as $T_{T,r}$.
Expert identifiers are extracted from a per-round trace in the vLLM fork and synchronized with CUDA-event timings; because of double buffering, covariates are shifted by one record.
The overhead of this instrumentation itself was not separately measured.

\subsection{Regression Visualization by $K$}
\label{sec:per-k-regression-figure}

\begin{figure*}[!t]
\centering
\includegraphics[width=\textwidth]{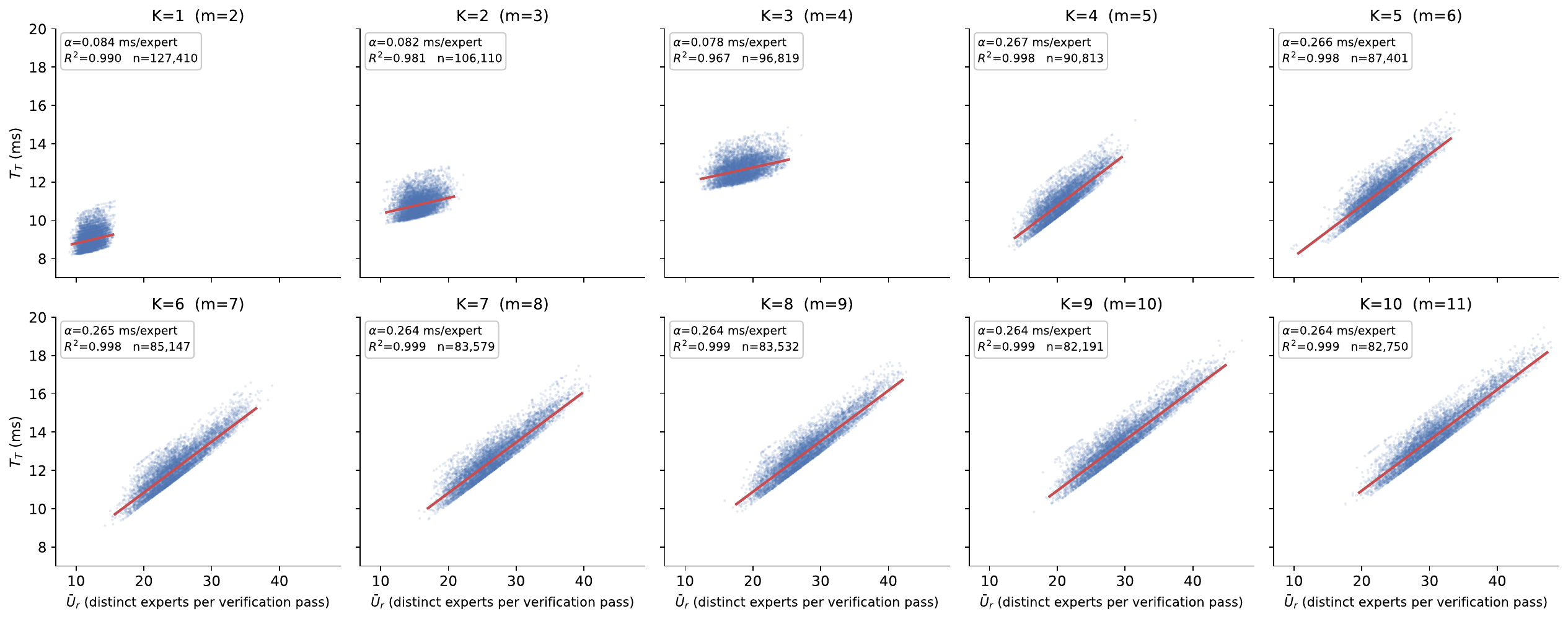}
\caption{$T_{T,r}$ against $\bar U_r$ for each $K$, with the fitted line and coefficients of model~\eqref{eq:target-cost}. The extended robustness protocol is given in Appendix~\ref{app:verification-cost-model}.}
\label{fig:cost-model-per-k-scatter}
\end{figure*}

Figure~\ref{fig:cost-model-per-k-scatter} shows the scatter of rounds $T_{T,r}$ against $\bar U_r$, together with the fitted line of model~\eqref{eq:target-cost}, separately for each $K\in\{1,\dots,10\}$. Going from $m=4$ to $m=5$, the slope $\alpha(m)$ increases roughly threefold, consistent with the breakpoint discussed above; the in-sample $R^2$ remains high across the whole range ($0.967$--$0.999$).

\subsection{Quantitative Validation and Scope of Applicability}
\label{sec:cost-model-validation}

The model was estimated on $925\,752$ rounds from a sweep over $K\in\{1,\dots,10\}$ for Qwen3-Coder-30B-A3B-Instruct and EAGLE-3 on a single NVIDIA A100 80GB.
Under GroupKFold grouped by generation, $\bar U_r$ is the strongest single predictor of verification time.
In the per-$K$ runs, the in-sample $R^2$ of model~\eqref{eq:target-cost} is $0.9669$--$0.9991$, and OOS MAPE is $0.26$--$0.59\%$.
Its standalone predictive power is uneven across domains: OOS $R^2$ is about $0.60$--$0.64$ for humanities, writing, and roleplay, versus $0.87$ for code\_math.

For the round's total cost, a separate draft-time model is used
\begin{equation}
\widehat T_D(K)=(0.459K+0.073)\,\mathrm{ms},
\label{eq:draft-cost-model}
\end{equation}
with CV MedAPE around $0.8\%$; this model supplies the draft term in the offline analysis.
The coefficients describe a predictive, not a causal, relationship, and only for the fixed model, backend, GPU, CUDA-graph mode, and $B=1$; the full protocol is given in Appendix~\ref{app:verification-cost-model}.

\section{The Sequence-Conditioned Offline Oracle (Methodology)}
\label{sec:offline-oracle}

As shown in the previous section, the verification-step time in MoE models is estimated by a linear model that accounts for prefix length and the number of unique activated experts. However, it is precisely the input-dependent routing --- how specific generated tokens activate different sets of experts --- that creates high variance in verification cost. Because of this instability, blindly fixing a speculative-batch size $k$ leads to suboptimal outcomes: the system inevitably risks spending compute on loading experts for tokens that will end up being rejected.

To find the absolute theoretical speedup limit and understand how the ideal strategy should balance token-acceptance probability against verification cost, we formulate the problem as a stochastic offline process over a known reference (ground-truth) sequence. This lets us build a hybrid Oracle whose behavior serves as a reference point for designing future adaptive online systems.

\subsection{Formulation as a Stochastic Shortest Path (SSP)}

For a rigorous search of the ideal speculation budget, we formulate the decoding process as a Stochastic Shortest Path (SSP) problem. The Markov Decision Process (MDP) formalization is as follows:

\begin{itemize}
    \item \textbf{State space ($S$):} The system's state is the current position $i$ in the reference (ground-truth) sequence $X_{1:N} = (x_1, x_2, \dots, x_N)$ of length $N$. Position $N$ acts as an absorbing state signaling the end of generation. Reaching this state terminates the process with zero residual cost ($dp[N] = 0$).

    \item \textbf{Action space ($A$):} While in state $i$, the Oracle takes an action --- it chooses the speculative-batch size (draft length) $k \in \{1 \dots L_{max}\}$.

    \item \textbf{Transitions:} Given a chosen batch size $k$, the draft model produces a hypothesis, of which the target model may accept $a$ tokens, $0 \le a \le k$. Because the speculative-decoding algorithm always computes one bonus token from the target model's logits, the system advances even under a complete hypothesis mismatch ($a = 0$). Thus, from position $i$ the system deterministically transitions to the new state $\min(i + a + 1, N)$.

    \item \textbf{Termination guarantee (Proper Policy):} A well-posed SSP problem requires the existence of at least one policy guaranteed to reach the absorbing state. Since the number of accepted tokens is always $a \ge 0$, every step advances the position by at least $1$. Hence, any strategy is guaranteed to reach the end of the sequence $N$ in at most $N$ steps. This property makes the transition graph a directed acyclic graph (DAG) and guarantees strict convergence of the Bellman problem.
\end{itemize}

\subsection{Counterfactual Simulation of Cost and Transitions}

Solving the SSP problem and finding the minimum expected generation time requires defining two components for each position $i \in \{0, \dots, N-1\}$ and speculation budget $k \in \{1, \dots, L_{max}\}$: the transition probabilities $P(a \mid k)$ and the verification cost estimate $Cost(i, k, a)$. Directly sampling trajectories online would cause a ``collapse of the future'', since incorrectly generated tokens would drift generation away from the reference text. To address this, we apply counterfactual simulation.

\paragraph{Transition probability distribution $P(a \mid k)$.}
For position $i$ and batch length $k$, the probability that the target model accepts exactly $a$ tokens ($0 \le a \le k$) is determined by the draft model's autoregressive confidence on the reference sequence $X_{1:N}$.

Let $p_j = P_{draft}(x_{i+1+j} \mid X_{\le i+j})$ denote the probability that the draft model correctly predicts the $(j+1)$-th token of the speculative window (i.e., the reference token $x_{i+1+j}$, given the prefix $X_{\le i+j}$, where $j \in \{0, \dots, k-1\}$). Then the probability of accepting the first $a$ tokens and rejecting the $(a+1)$-th token (for $a < k$) is computed as:
\begin{equation}
P(a \mid k) = \left( \prod_{j=0}^{a-1} p_j \right) \cdot (1 - p_a)
\end{equation}
For the case of full hypothesis acceptance ($a = k$), the probability is $P(k \mid k) = \prod_{j=0}^{k-1} p_j$.

\paragraph{Counterfactual simulation of ``junk tails'' and cost estimation.}
The verification cost $Cost(i, k, a)$ is determined not only by the number of accepted tokens but also by the structure of experts activated by the entire candidate chain of length $k$. To estimate this cost without disturbing the main state $i$, we model the physics of a speculative error by sampling hypothetical suffixes (``junk tails''):
\begin{enumerate}
    \item For each $a \in \{0, \dots, k\}$, a candidate chain $\tilde{X}^{(a)}$ of length $k$ is formed. Its first $a$ tokens exactly match the ground truth ($x_{i+1}, \dots, x_{i+a}$).
    \item At position $a$, an error is artificially induced: the token $\tilde{x}_{a}$ is sampled from the draft model's distribution with the reference token's probability $x_{i+1+a}$ zeroed out. The remaining $k - a - 1$ tokens are sampled freely.
    \item The resulting suffix batch $\tilde{X}^{(a)}$ is fed to the target model using the previously cached KV cache for prefix $X_{\le i}$. The number of unique activated experts $U(i, k, a)$ is extracted from the MoE layers' router logits.
    \item Converting the expert count $U(i, k, a)$ into the final step time is done through the cost model $Cost(i, k, a) = t_{draft}(k) + t_{target}(U, L_{prefix})$, where $L_{prefix} = L_{prompt} + i$.
\end{enumerate}

The simulated suffix $\tilde{X}^{(a)}$ is used solely to measure the expert count $U(i, k, a)$ and is then discarded. For computational efficiency, the Oracle uses a single-sample Monte Carlo estimate of the junk tail, relying on the local variance in activated experts averaging out over the length of the whole sequence. In this way, the Oracle correctly accounts for the routing overhead of incorrectly predicted tokens while keeping the state space strictly on the linear ground-truth axis.

\subsection{The Bellman Oracle and Speedup Bounds}

Given the probability distribution ${P(a \mid k)}$ and the precomputed cost matrix $Cost(i, k, a)$, the minimum expected generation time for the remaining sequence (from position $i$ to the end $N$) is given by the Bellman equation:
\begin{equation}
\begin{split}
dp[i] &= \min_{k \in \{1 \dots L_{max}\}} \Bigg[ \sum_{a=0}^{k} {P(a \mid k)} \cdot \\
&\quad \Big( Cost(i, k, a) + dp[\min(i + a + 1, N)] \Big) \Bigg]
\end{split}
\end{equation}

Because every speculative step deterministically advances the position by at least one token, the state graph is strictly acyclic (a DAG). This allows the final expectation $dp[0]$ to be computed in a single pass of standard backward induction from the absorbing state $dp[N] = 0$. The array of $k$ values that minimize the equation at each step $i$ forms our Oracle's optimal offline policy.

It is important to stress the limits of this estimate's applicability. Because the counterfactual simulation (Section 4.2) is rigidly tied to the reference sequence $X_{1:N}$ and ignores alternative valid generation paths, the resulting value $dp[0]$ and its corresponding speedup $S = Cost_{AR} / dp[0]$ represent an efficiency limit only for reproducing this specific text. Nevertheless, for the sequence-preserving speculative-decoding problem, this estimate provides a reliable analytical reference point (an upper bound) against which any online heuristic lacking access to future context can be compared.

\section{Analysis of the Oracle's Behavior}

To empirically evaluate the developed Oracle, we used the Math Reasoning subset (80 questions) of the SpecBench benchmark. Qwen3-Coder-30B-A3B-Instruct served as the target model, and SGLang-EAGLE3-Qwen3-Coder-30B-A3B-Instruct-SpecForge served as the draft model. All simulations and profiling measurements were carried out on a single NVIDIA A100 80GB GPU.

\subsection{The Limits of Speculation (Quantitative Bounds)}

We first examine the algorithm's macro-behavior as the maximum allowed speculation budget $L_{max}$ increases.

% --- FIGURE 1 BLOCK (single column) ---
\begin{figure}[t]
\centering
\includegraphics[width=\linewidth]{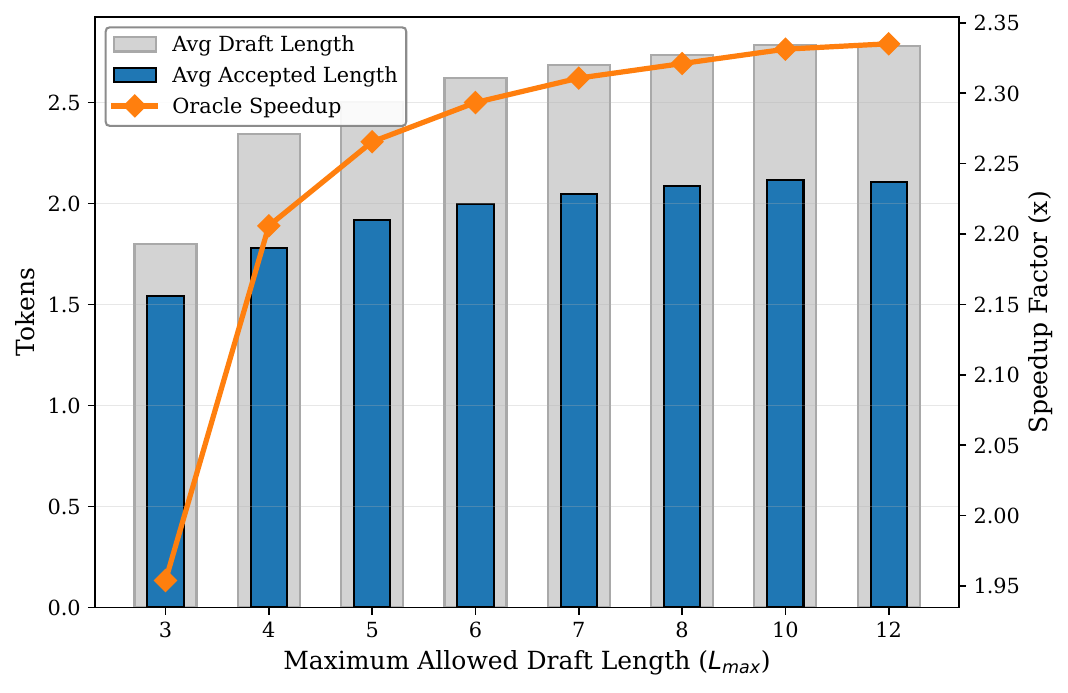}
\caption{\textbf{Quantitative bounds of the MoE Stochastic Oracle under varying speculation budgets $L_{max}$.} As $L_{max}$ increases, Theoretical Maximum Speedup ($S_{max}$), Mean $k$, and Mean Accepted Length exhibit diminishing returns and saturation. The Oracle selectively caps $k$ to avoid costly MoE expert activations on low-probability draft tokens.}
\label{fig:quantitative_bounds}
\end{figure}
% --------------------------------------

Figure~\ref{fig:quantitative_bounds} shows Oracle Speedup, Avg Draft Length, and Avg Accepted Length as functions of $L_{max}$. The main result of this experiment is a strong diminishing-returns effect. As the plot shows, beyond $L_{max} = 6$ the curves essentially stop growing: the final speedup plateaus at roughly 2.34x, and the average accepted-batch length (Avg Accepted Length) saturates at $\approx 2.1$ tokens. Most importantly, the Oracle forcibly caps the average generated draft length (Avg Draft Length) at $\approx 2.8$, effectively ignoring the larger budgets available up to $L_{max} = 12$.

This dynamic is explained by the very mechanics of the DP algorithm's decision-making. As $L_{max}$ grows, the Oracle gains the ability to use long speculative chains, yet it deliberately declines to use it. The reason lies in the strict accounting of acceptance probability: the draft model's predictive power inevitably declines for farther-out tokens. Because in the MoE architecture processing each rejected token carries a noticeable penalty from loading unique experts, the Oracle balances the falling success probability $P(a \mid k)$ against the rising cost in the matrix $Cost(i, k, a)$. As a result, the algorithm prefers to settle on shorter but more reliable chains, forcibly stabilizing the average value of $k$ regardless of how large a budget $L_{max}$ it is formally allowed.

\subsection{Delta Space Analysis and the Linear Boundary}

To understand the logic by which the Oracle truncates speculative-batch length, we examined the space of marginal decisions (Delta Space) at the local level. For each generation step, the Oracle selects the optimal draft length $k_{opt}$. All other possible lengths $k \neq k_{opt}$ are treated as rejected alternatives.

For each such alternative, we computed two marginal deltas relative to the Oracle's choice:
\begin{itemize}
    \item $\Delta \mathbb{E}[a] = \mathbb{E}[a]_k - \mathbb{E}[a]_{k_{opt}}$ --- the marginal difference in the expected number of accepted tokens;
    \item $\Delta \mathbb{E}[Cost] = \mathbb{E}[Cost]_k - \mathbb{E}[Cost]_{k_{opt}}$ --- the marginal difference in expected cost (number of activated experts).
\end{itemize}

% --- FIGURE 2 BLOCK (single column) ---
\begin{figure}[t]
\centering
\includegraphics[width=\linewidth]{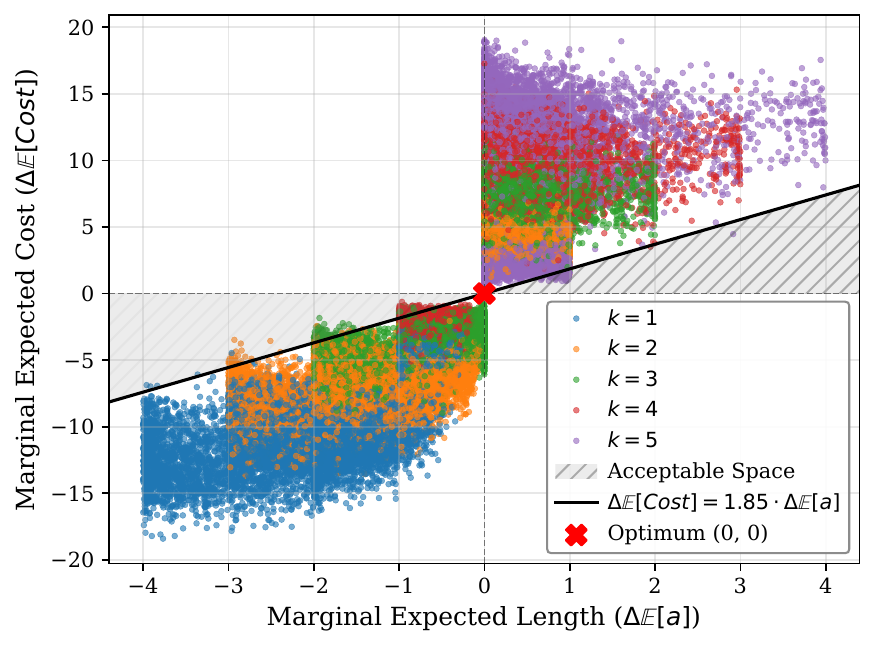}
\caption{\textbf{Delta Space Analysis ($\Delta \mathbb{E}[\text{Cost}]$ vs. $\Delta \mathbb{E}[a]$) of rejected draft candidates relative to the Oracle's choice $(0,0)$.} Quadrants II and IV are strictly empty, proving strict monotonicity. Candidates in Quadrant I (risk zone) and Quadrant III (caution zone) form a linear boundary passing through the origin.}
\label{fig:delta_space}
\end{figure}
% ------------------------------------

If the Oracle's choice is placed at the origin $(0,0)$, the rejected candidates form a dense cloud of points around it (see Figure~\ref{fig:delta_space}). Analysis of the distribution of these points reveals three fundamental behavioral patterns:

\paragraph{1. Structural monotonicity (empty quadrants II and IV).}
The upper-left ($\Delta Cost > 0, \Delta a < 0$) and lower-right ($\Delta Cost < 0, \Delta a > 0$) quadrants of the plot are, as expected, completely empty. This is a direct consequence of the algorithm's construction and the cost model: generating additional tokens structurally requires additional computation steps. Thus, the impossibility of increasing $\mathbb{E}[a]$ while decreasing $\mathbb{E}[Cost]$ serves as a natural \textit{sanity check} on the correctness of the constructed space of marginal deltas.

\paragraph{2. The risk and caution thresholds (quadrants I and III).}
The upper-right quadrant (the risk zone) concentrates batches that were too long: they promised more tokens ($\Delta a > 0$) but were rejected for a disproportionately high expert overpayment ($\Delta Cost > 0$). The lower-left quadrant (the caution zone) contains chains that were too short: they saved time ($\Delta Cost < 0$) but were rejected for a substantial loss of potential tokens.

\paragraph{3. The linear boundary.}
The most important observation is that the edges of the rejected-point clouds in quadrants I and III line up along a single line passing through the origin. This line (constructed from the 1st percentile of the ratio $\Delta Cost / \Delta a$) forms the boundary of the infeasible region. No rejected point can physically cross this line, since otherwise the algorithm would have recognized it as the optimal decision.

The existence of this constant linear boundary has substantial practical significance. It demonstrates that the globally optimal search over the Bellman equation, which requires knowledge of the entire future text, reduces at the local level to maintaining a simple balance: the ratio $\frac{\Delta \mathbb{E}[Cost]}{\Delta \mathbb{E}[a]}$. This means that building effective online speculative-decoding algorithms for MoE does not require heavy predictive models. It suffices to estimate the marginal cost of adding the next token to the chain. If this ratio breaks through the constant slope barrier, the algorithm should immediately halt draft generation (early exit) and proceed to verification.

\section{Conclusion}
In this work, we studied the problem of input-dependent verification cost in speculative decoding for MoE models. To analyze the trade-off between token-acceptance probability and expert-activation cost, we developed a diagnostic offline method. By formulating the problem as a stochastic shortest path (SSP) over a fixed reference sequence and applying counterfactual simulation of ``junk tails'', we built a Sequence-Conditioned Oracle that provides an upper-bound speedup estimate for a given text.

Analysis of the Oracle's offline decisions yields two main conclusions:
\begin{itemize}
    \item \textbf{Diminishing returns:} As the available budget $L_{max}$ grows, the final speedup saturates quickly. The Oracle deliberately limits the average speculative-batch length, since the routing overhead of low-probability tokens begins to outweigh the benefit of their possible acceptance.
    \item \textbf{A linear boundary in delta space:} Examining the rejected marginal alternatives shows that the Oracle's choice forms a strict boundary. Locally optimal decisions obey a necessary condition: the ratio of marginal cost to the expected gain in accepted tokens ($\frac{\Delta \mathbb{E}[Cost]}{\Delta \mathbb{E}[a]}$) must not exceed a constant threshold.
\end{itemize}

The marginal structure we uncover demonstrates that decisions requiring knowledge of the entire sequence have a clear local interpretation. This observation does not offer a ready-made system, but it motivates the design of future online heuristics that could use the marginal-cost ratio for early exit from speculative generation.

\section{Limitations}
\label{sec:limitations}

This study makes a number of assumptions that bound the applicability of the proposed offline method:

\begin{itemize}
    \item \textbf{Batch size and configuration.} The analysis is confined to the \texttt{Qwen3-Coder-30B-A3B-Instruct} and \texttt{EAGLE-3} (SpecForge) pairing at batch size $B=1$. Generation at $B>1$ introduces inter-sequence expert sharing, which fundamentally changes the structure of the overhead and requires a different cost model.
    \item \textbf{Domain effect.} The evaluation was carried out on the Math Reasoning subset. This domain is aligned with the draft model's training distribution, which yields high baseline confidence ($p_j$) and allows the MoE-routing penalties to be isolated honestly. While the diminishing-returns effect itself is general, the exact budget saturation point ($L_{max}$) will vary for text with different entropy.
    \item \textbf{Oracle approximations.} For computational efficiency, transition probabilities are estimated via draft confidence (surrogate probability), and the cost of rejected tokens in the matrix $Cost(i, k, a)$ is computed via a single-shot Monte Carlo sampling of suffixes, which introduces local noise into the point estimates.
    \item \textbf{No online policy.} This work establishes the existence of a linear boundary in delta space as a necessary condition for optimality. However, designing and end-to-end evaluating an adaptive online heuristic (early exit) that exploits this rule is left for future work.
\end{itemize}

\bibliography{references}

\clearpage
\appendix
\setcounter{section}{0}
\setcounter{table}{0}
\setcounter{figure}{0}
\setcounter{equation}{0}
\renewcommand{\thetable}{\Alph{section}\arabic{table}}
\renewcommand{\thefigure}{\Alph{section}\arabic{figure}}
\renewcommand{\theequation}{\Alph{section}\arabic{equation}}
\renewcommand{\theHequation}{\Alph{section}.\arabic{equation}}
\input{appendix_verification_cost_en}

\FloatBarrier
\setcounter{table}{0}
\setcounter{figure}{0}
\setcounter{equation}{0}
\input{appendix_linguistic_analysis_en}

\end{document}

%% file: appendix_verification_cost_en.tex
\section{Verification Cost Factors and the Predictive Model for \texorpdfstring{$T_T$}{TT}}
\label{app:verification-cost-model}

\subsection{Research Question and Hypotheses}

The central thesis of this analysis is that for MoE targets, verification cost can be a bottleneck of speculative decoding and is linked to the activation of unique experts.
Three testable hypotheses follow from it:
\begin{enumerate}
    \item[\textbf{H1.}] The routing signal $\bar U_r$ --- the number of unique experts touched by the verification pass --- predicts round latency $T_T$ more accurately than each of the five other factors considered.
    \item[\textbf{H2.}] Within a fixed system regime, one additional unique expert is associated with an approximately constant increment in latency.
    \item[\textbf{H3.}] As the budget $K$ grows, the predictive association of $\bar U_r$ with $T_T$ strengthens relative to the association with context length $L$.
\end{enumerate}

What follows describes the experimental setup and data, candidate factors, the protocol and results of the factor analysis, the specification and fit of the final model, the budget as a model parameter, the draft-time model, and robustness checks.

\subsection{Experimental Setup}

All runs were executed on a single NVIDIA A100 80GB GPU in bfloat16.
Stage timings were captured with CUDA events inside a vLLM fork with per-round tracing; the full sweep over $K\in\{1,\dots,10\}$ took about nine GPU-hours.

\begin{table}[!tbp]
\centering
\footnotesize
\begin{tabularx}{\columnwidth}{@{}p{0.24\columnwidth}X@{}}
\toprule
Parameter & Value \\
\midrule
Target & Qwen/Qwen3-Coder-30B-A3B-Instruct; MoE with 128 experts and top-$k_e=8$ routing \\
Draft head & lmsys/SGLang-EAGLE3-Qwen3-Coder-30B-A3B-Instruct-SpecForge \\
GPU and stack & NVIDIA A100 80GB; vLLM fork with per-round trace; FULL CUDA graphs \\
Mode & $B=1$, synchronous scheduling \\
Sampling & $T=0.7$, \texttt{top\_p}$=1.0$, seed $=0$ \\
Other & bfloat16, \texttt{max\_model\_len}$=4096$, \texttt{gpu\_mem\_util}$=0.85$, \texttt{max\_new\_tokens}$=1000$ \\
Dataset & SpecBench, 560 questions, 6 domains \\
Sweep & $K\in\{1,\dots,10\}$, one full run for each $K$; about nine GPU-hours \\
Timing & Per-stage CUDA events; double-buffer alignment via a one-record covariate shift \\
Volume & $925\,752$ speculative rounds; 641 generations per run \\
\bottomrule
\end{tabularx}
\end{table}

The unit of observation is the speculative round.
Each record contains the stage timings $T_T$ and $T_D$, the verification-microbatch size $m=K+1$, the routing signal $\bar U_r$, the context length $L$, KV-cache occupancy, and the CUDA-graph bucket.
Records from the prefill phase were excluded from estimation but were used to mark generation boundaries.
The generation identifier defined the clusters for robust standard errors and the grouping used in cross-validation.
Rounds are not independent experimental units: they are nested within generations, and the same questions recur across different $K$.
GroupKFold limits train/test leakage within each analysis but does not eliminate dependence across the $K$-slices of a single question.

\subsection{Measurement Conventions}

The definitions of $K$, $m=K+1$, useful progress, $T_D$, and $T_T$ are given in the main text; here we fix two boundaries of the recorded timings.
A separate quantity $T_{\mathrm{reject},r}$ would cover rejection sampling and post-processing, which are not included in the recorded $T_{D,r}$ and $T_{T,r}$; there is no separate measurement of this overhead in the trace.
Rejected positions are nonetheless part of the target model's microbatch and are therefore counted in $T_{T,r}$ and $\bar U_r$.

The extra target-model position increases useful progress only when it becomes an emitted output token.
If this position is not emitted because the sequence ends, its computation remains part of $T_{T,r}$ but does not increase $G_r$.

\begin{table*}[!t]
\centering
\small
\resizebox{\textwidth}{!}{%
\begin{tabular}{@{}rrrrrr@{}}
\toprule
$K$ & $m$ & Rounds & Median $T_T$, ms & Median $\bar U_r$ & Median $L$ \\
\midrule
1  & 2  & 127410 & 8.98  & 12.1 & 605 \\
2  & 3  & 106110 & 10.81 & 15.5 & 606 \\
3  & 4  & 96819  & 12.63 & 18.2 & 605 \\
4  & 5  & 90813  & 10.94 & 20.6 & 605 \\
5  & 6  & 87401  & 11.51 & 22.6 & 600 \\
6  & 7  & 85147  & 12.07 & 24.5 & 591 \\
7  & 8  & 83579  & 12.48 & 26.1 & 586 \\
8  & 9  & 83532  & 12.91 & 27.5 & 594 \\
9  & 10 & 82191  & 13.35 & 29.0 & 589 \\
10 & 11 & 82750  & 13.67 & 30.2 & 594 \\
\bottomrule
\end{tabular}}
\caption{Dataset summary by run. Median $L$ ranges between 586 and 606 tokens. The non-monotonicity of $T_T$ at $K=3\rightarrow4$ coincides with a change in the estimated $\beta_0$; without a randomized repeat sweep it cannot be causally attributed to a CUDA-graph mode.}
\label{tab:appendix-run-summary}
\end{table*}

% compact appendix layout: barrier removed
\subsection{Candidate Factors}

Six measurable candidate factors were considered; each corresponds to a physical hypothesis about its mechanism of association with $T_T$.
The speculation budget $K$ is the only operator-controlled quantity.
In the analysis it is represented by the number of positions in the verification pass, $m=K+1$; within a single run, $m$ is deterministically set by the budget, so a separate factor $K$ is redundant.

\begin{table*}[!t]
\centering
\small
\begin{tabularx}{\textwidth}{@{}p{0.16\textwidth}p{0.37\textwidth}X@{}}
\toprule
Factor & What it measures & Mechanistic hypothesis \\
\midrule
$\bar U_r$ & Number of unique experts per pass, averaged across layers & Expert-weight traffic from HBM \\
$L$ & Total context, tokens & KV-cache read and attention \\
$m$ & Number of positions in the verification pass & Microbatch and CUDA-graph size \\
$E_{\max}$ & Maximum number of tokens per single expert & Load imbalance \\
KV usage & KV-cache occupancy & Allocator pressure \\
Graph bucket & Size of the captured graph & Padding; at $B=1$ the bucket is equivalent to $m$ \\
\bottomrule
\end{tabularx}
\end{table*}

\subsubsection{Unique Expert Counting}

Let $G_\ell$ denote the router's scoring function in layer $\ell$, let $k_e$ denote the number of experts selected per token, and let $\mathcal L_{\mathrm{MoE}}$ denote the set of MoE layers of the target model.
For batch element $b$, verification position $t$, and token representation $x_{\ell,b,t}$, the local set of selected experts is
\begin{equation}
\mathcal E_{\ell,b,t}
=
\operatorname{TopK}(G_\ell(x_{\ell,b,t}),k_e).
\label{eq:token-local-experts}
\end{equation}
Here $k_e$ differs from $K$: the former sets the number of experts per token, the latter the number of draft positions.

For round $r$, the set $\mathcal I_r$ contains all pairs $(b,t)$ in the verification microbatch, including the extra target-model position.
The union of experts in layer $\ell$ equals
\begin{equation}
\mathcal U_{\ell,r}
=
\bigcup_{(b,t)\in\mathcal I_r}\mathcal E_{\ell,b,t}.
\label{eq:microbatch-expert-union}
\end{equation}
The signal $\bar U_r$ is defined as the mean of $|\mathcal U_{\ell,r}|$ across MoE layers, per Eq.~\eqref{eq:layer-averaged-experts}.

% compact appendix layout: barrier removed
\subsection{Factor Importance}

Factor importance was determined by its predictive power on held-out generations.
Rounds were split into five folds grouped by generation: all rounds of a given response fell into the same fold.
For each fold, an OLS model was fit on the remaining four, after which out-of-sample $R^2$ and RMSE were computed on the held-out fold.
Standalone factor strength was given by the across-fold mean with a 95\% $t$-interval at four degrees of freedom.
The incremental contribution beyond $\bar U_r$ was measured by the paired RMSE reduction of the model $\bar U_r+$factor relative to the $\bar U_r$-only model.

For correlated factors, it is precisely the paired incremental estimate that is meaningful: a factor may predict well on its own by moving together with $\bar U_r$ while carrying no independent information.

\begin{table*}[!tbp]
\centering
\small
\resizebox{\textwidth}{!}{%
\begin{tabular}{@{}lrrrr@{}}
\toprule
Factor & OOS $R^2$, alone & OOS RMSE, ms & $\Delta$RMSE beyond $\bar U_r$, ms & corr with $\bar U_r$ \\
\midrule
$\bar U_r$ & $0.787\pm0.026$ & $0.831\pm0.037$ & --- & 1.00 \\
$m=K+1$ & $0.515\pm0.033$ & $1.256\pm0.020$ & $+0.014\pm0.003$ & 0.86 \\
Graph bucket & $0.515\pm0.033$ & $1.256\pm0.020$ & $+0.014\pm0.003$ & 0.86 \\
$E_{\max}$ & $0.362\pm0.029$ & $1.442\pm0.014$ & $+0.015\pm0.002$ & 0.75 \\
KV usage & $0.077\pm0.017$ & $1.734\pm0.051$ & $+0.139\pm0.007$ & 0.03 \\
$L$ & $0.079\pm0.016$ & $1.732\pm0.051$ & $+0.146\pm0.008$ & 0.03 \\
\bottomrule
\end{tabular}}
\caption{Descriptive predictive importance of factors on the pool of $925\,752$ rounds; intervals are 95\% CIs across folds. $\bar U_r$ is the strongest single predictor, and $L$ gives the largest additional error reduction on top of it.}
\label{tab:appendix-factor-importance}
\end{table*}

Three observations follow from Table~\ref{tab:appendix-factor-importance}.
First, $\bar U_r$ is the strongest of the single predictors considered, which supports the predictive version of H1 in the studied environment.
Second, the budget predicts latency moderately on its own, $R^2\approx0.52$, but on top of $\bar U_r$ it reduces error by only $0.014\pm0.003$ ms.
The budget's predictive association with cost is mostly absorbed by its covariation with $\bar U_r$; the residual differences show up in $\alpha(m)$ and $\beta_0(m)$.
Third, $L$ is weak on its own but is almost uncorrelated with $\bar U_r$ and gives an additional RMSE reduction of $0.146\pm0.008$ ms, which is consistent with, but does not identify, a separate KV-cache cost channel.
$L$ and KV usage have nearly identical standalone accuracy and may encode overlapping properties of context length and memory state.
A joint VIF diagnostic or a separate ablation of their redundancy was not performed; the choice of $L$ is based on interpretability and predictive contribution, not on identifying a unique causal channel.

The standardized association of $T_T$ with $\bar U_r$ within a single run grows monotonically from $0.20\sigma$ at $K=1$ to $0.93\sigma$ at $K=10$, while the association with $L$ declines from $0.96\sigma$ to $0.31\sigma$; the width of the 95\% CIs does not exceed $\pm0.004$.
At small microbatches the standardized association of $T_T$ is stronger with $L$; at large microbatches it is stronger with $\bar U_r$; this is descriptively consistent with H3.

% compact appendix layout: barrier removed
\subsubsection{Breakdown by Dataset Category}

Within each category, the same GroupKFold-by-generation protocol was applied, with 95\% intervals across folds.
The comparisons across the fourteen categories are descriptive; the intervals are not adjusted for multiplicity and are not used as a family of confirmatory tests.

\begin{table*}[!tbp]
\centering
\scriptsize
\resizebox{\textwidth}{!}{%
\begin{tabular}{@{}lrrrrr@{}}
\toprule
Category & Rounds & OOS $R^2$: $\bar U_r$ & OOS $R^2$: $\bar U_r+L$ & $\Delta$RMSE from $L$, ms & Final $R^2$ \\
\midrule
code\_math & 326429 & $0.870\pm0.023$ & $0.903\pm0.018$ & $+0.098\pm0.007$ & 0.9993 \\
extraction & 5839 & $0.861\pm0.097$ & $0.864\pm0.090$ & $+0.005\pm0.019$ & 0.9991 \\
math\_reasoning & 57788 & $0.843\pm0.039$ & $0.851\pm0.033$ & $+0.015\pm0.014$ & 0.9990 \\
coding & 47116 & $0.809\pm0.079$ & $0.904\pm0.064$ & $+0.281\pm0.081$ & 0.9991 \\
translation & 16594 & $0.797\pm0.023$ & $0.795\pm0.022$ & $-0.002\pm0.008$ & 0.9991 \\
rag & 63141 & $0.766\pm0.011$ & $0.774\pm0.007$ & $+0.012\pm0.024$ & 0.9982 \\
qa & 75787 & $0.742\pm0.051$ & $0.746\pm0.048$ & $+0.006\pm0.005$ & 0.9987 \\
math & 23111 & $0.741\pm0.061$ & $0.794\pm0.088$ & $+0.097\pm0.144$ & 0.9987 \\
stem & 42275 & $0.704\pm0.086$ & $0.757\pm0.055$ & $+0.074\pm0.093$ & 0.9986 \\
reasoning & 24823 & $0.685\pm0.172$ & $0.756\pm0.102$ & $+0.095\pm0.092$ & 0.9986 \\
summarization & 106868 & $0.660\pm0.028$ & $0.747\pm0.012$ & $+0.114\pm0.033$ & 0.9983 \\
roleplay & 33854 & $0.641\pm0.084$ & $0.684\pm0.086$ & $+0.055\pm0.088$ & 0.9986 \\
writing & 34043 & $0.603\pm0.167$ & $0.687\pm0.182$ & $+0.100\pm0.130$ & 0.9986 \\
humanities & 67818 & $0.602\pm0.125$ & $0.745\pm0.078$ & $+0.182\pm0.089$ & 0.9987 \\
\midrule
All categories & 925752 & $0.787\pm0.026$ & 0.856 & $+0.146\pm0.008$ & 0.999 \\
\bottomrule
\end{tabular}}
\caption{Factor importance by category, sorted by the OOS $R^2$ of the $\bar U_r$-only model. The summary row corresponds to the full pool.}
\label{tab:appendix-factor-categories}
\end{table*}

\begin{figure*}[!tbp]
\centering
\includegraphics[width=0.80\textwidth]{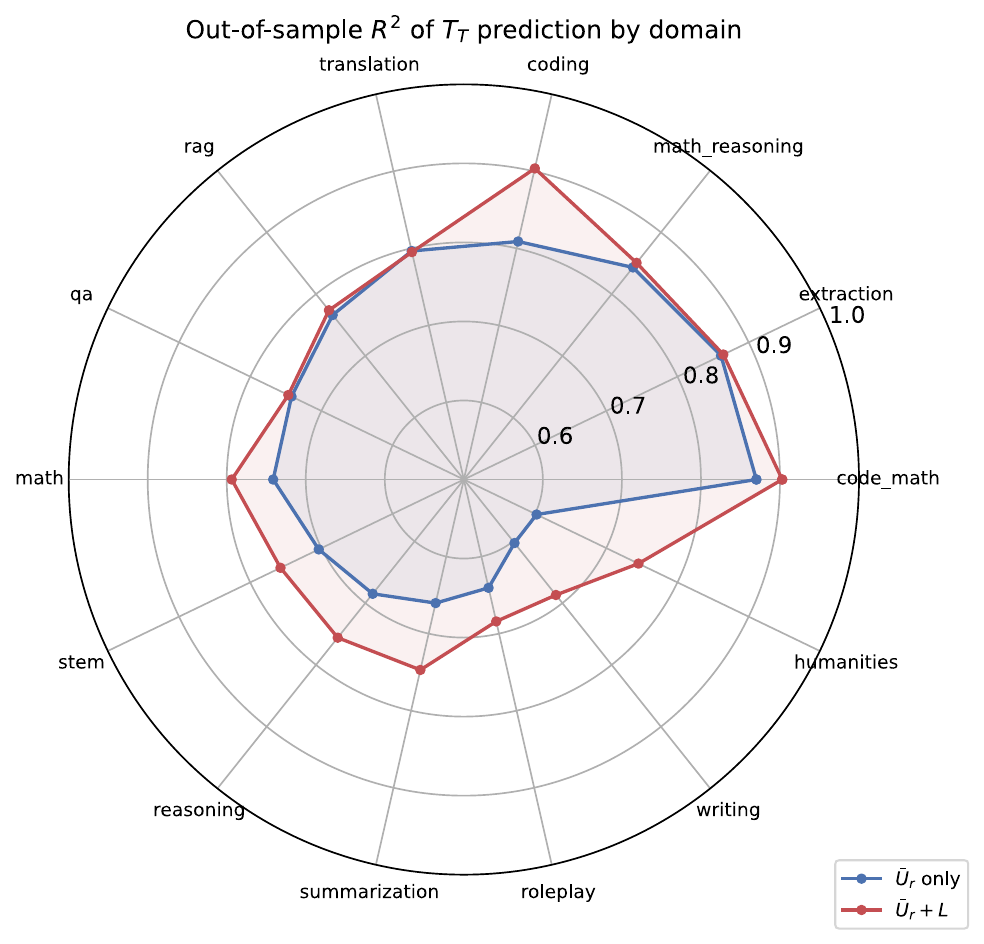}
\caption{Out-of-sample $R^2$ for predicting $T_T$ by category: $\bar U_r$ alone versus $\bar U_r+L$.}
\label{fig:appendix-factor-radar}
\end{figure*}

In the descriptive comparison, $\bar U_r$ remains the leading single predictor across all fourteen categories, while the additional contribution of $L$ is uneven and in places has wide intervals.
The share of variance explained by $\bar U_r$ alone ranges from 0.60 for humanities and writing to 0.87 for code\_math.
In structured domains $\bar U_r$ explains a larger share of variation, whereas in free-form text $L$ provides a markedly larger additional predictive contribution.
The wide intervals for small subcategories reflect their small sample size.

% compact appendix layout: barrier removed
\subsection{Predictive Model for \texorpdfstring{$T_T$}{TT}}

We use the specification from Eq.~\eqref{eq:target-cost}: the budget enters categorically through free $\beta_0(m)$ and $\alpha(m)$ for each $m=K+1$, while the context coefficient $c$ is shared.

The choice of parameters is motivated as follows.
$\bar U_r$ is an observed cost predictor consistent with reading expert weights from HBM.
A rough bandwidth-based estimate, $48$ layers of $9.4$ MB each, or about $0.45$ GB at $2$ TB/s, gives about $0.23$ ms, matching the order of magnitude of the estimated $\alpha$ at $m\geq5$.
Context $L$ gives the largest additional reduction in predictive error on top of $\bar U_r$ and is consistent with the KV-cache read channel.
The categorical form describes the two observed ranges of $\alpha(m)$ and the shift in $\beta_0(m)$ between $m=4$ and $m=5$.
The split point was chosen after inspecting the full dataset; the subsequent Group\-KFold evaluation estimates coefficients under an already-fixed form and is not a prospective test of the changepoint or its cause.

Parameters were estimated by OLS separately on the subsample of each $K$, under the specification $T_T\sim\bar U_r+L$.
Standard errors were clustered by generation; out-of-sample metrics were computed via GroupKFold(5) with the same grouping.

\begin{figure*}[!tbp]
\centering
\includegraphics[width=\textwidth]{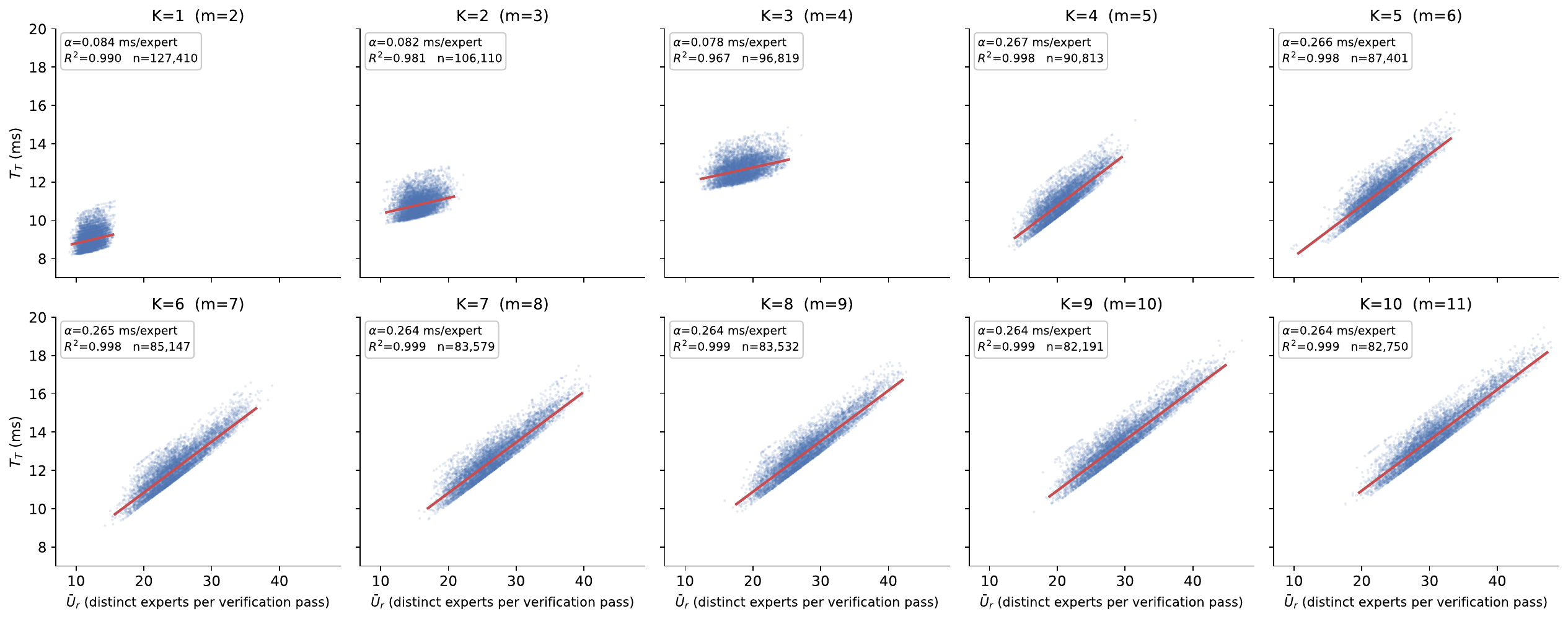}
\caption{Round scatter and fitted line for each of the ten values of $K$ on shared axes. As $K$ grows, the cloud shifts to the right, and at the $m=4/5$ boundary the slope increases roughly threefold.}
\label{fig:appendix-per-k-scatter}
\end{figure*}

\begin{table*}[!tbp]
\centering
\scriptsize
\resizebox{\textwidth}{!}{%
\begin{tabular}{@{}rrrrrrr@{}}
\toprule
$K$ & $\alpha$, ms/expert & $c$, $10^{-3}$ ms/token & $\beta_0$, ms & $N$ & $R^2$ & OOS MAPE \\
\midrule
1  & 0.0839 (0.0003) & 1.098 (0.001) & 7.294 (0.003)  & 127410 & 0.9896 & 0.44\% \\
2  & 0.0824 (0.0003) & 1.101 (0.002) & 8.854 (0.005)  & 106110 & 0.9806 & 0.51\% \\
3  & 0.0779 (0.0004) & 1.106 (0.002) & 10.532 (0.008) & 96819  & 0.9669 & 0.59\% \\
4  & 0.2675 (0.0002) & 1.099 (0.001) & 4.739 (0.003)  & 90813  & 0.9977 & 0.33\% \\
5  & 0.2664 (0.0003) & 1.103 (0.001) & 4.780 (0.006)  & 87401  & 0.9982 & 0.32\% \\
6  & 0.2654 (0.0001) & 1.106 (0.001) & 4.881 (0.003)  & 85147  & 0.9985 & 0.30\% \\
7  & 0.2642 (0.0001) & 1.107 (0.001) & 4.886 (0.002)  & 83579  & 0.9987 & 0.29\% \\
8  & 0.2643 (0.0001) & 1.102 (0.001) & 4.938 (0.002)  & 83532  & 0.9989 & 0.28\% \\
9  & 0.2638 (0.0001) & 1.105 (0.001) & 5.017 (0.002)  & 82191  & 0.9990 & 0.27\% \\
10 & 0.2638 (0.0001) & 1.106 (0.001) & 5.009 (0.002)  & 82750  & 0.9991 & 0.26\% \\
\bottomrule
\end{tabular}}
\caption{OLS estimates of the model $T_T=\beta_0+\alpha\bar U_r+cL$ by budget $K$. Parentheses give clustered standard errors; each run has 641 cluster-generations. Under these standard errors, all reported coefficients have $p<0.001$; because of the large number of nested rounds, the emphasis is on effect sizes, intervals, and OOS error. OOS MAPE is measured on held-out generation groups across five folds.}
\label{tab:appendix-per-k-model}
\end{table*}

In the observed sweep, $\alpha(m)$ is $0.078$--$0.084$ ms for $m\leq4$ and $0.264$--$0.268$ ms for $m\geq5$, i.e., the ratio of characteristic levels is about $3.2$.
In the small-$m$ range, $\beta_0$ grows from 7.3 to 10.5 ms, after which it shifts to 4.7 ms between adjacent configurations.
These differences describe the runs but, without interleaved repetition, do not identify a hardware cause for the boundary.

The coefficient $\alpha$ is interpreted as the predictive increment associated with one additional unique expert; $c$ as the increment associated with a context token, about 1.1 ms per 1000 tokens; $\beta_0$ absorbs the baseline launch cost of dense sublayers and the router.
This interpretation is consistent with the measurements, but the regression itself does not establish causality.

% compact appendix layout: barrier removed
\subsubsection{Model by Category}

The same regression was estimated within each of the fourteen categories under the specification $T_T\sim\bar U_r\times C(K)+L$.
Table~\ref{tab:appendix-model-categories} reports the slope at $K=8$ and the shared context coefficient.

\begin{table*}[!tbp]
\centering
\scriptsize
\resizebox{\textwidth}{!}{%
\begin{tabular}{@{}lrrrr@{}}
\toprule
Category & $\alpha$ at $K=8$ & $c$, $10^{-3}$ & $N$ & $R^2$ \\
\midrule
coding & 0.2625 (0.0005) & 1.110 (0.001) & 47116 & 0.9991 \\
code\_math & 0.2640 (0.0001) & 1.096 (0.001) & 326429 & 0.9993 \\
qa & 0.2643 (0.0004) & 1.165 (0.010) & 75787 & 0.9987 \\
stem & 0.2644 (0.0003) & 1.094 (0.002) & 42275 & 0.9986 \\
math\_reasoning & 0.2644 (0.0005) & 1.109 (0.003) & 57788 & 0.9989 \\
translation & 0.2647 (0.0005) & 1.190 (0.041) & 16594 & 0.9991 \\
extraction & 0.2647 (0.0006) & 1.051 (0.008) & 5839 & 0.9990 \\
reasoning & 0.2647 (0.0006) & 1.105 (0.004) & 24823 & 0.9985 \\
writing & 0.2649 (0.0004) & 1.098 (0.002) & 34043 & 0.9985 \\
math & 0.2649 (0.0004) & 1.099 (0.004) & 23111 & 0.9987 \\
rag & 0.2654 (0.0004) & 1.053 (0.007) & 63141 & 0.9981 \\
humanities & 0.2655 (0.0004) & 1.104 (0.001) & 67818 & 0.9986 \\
summarization & 0.2660 (0.0002) & 1.114 (0.001) & 106868 & 0.9983 \\
roleplay & 0.2664 (0.0006) & 1.085 (0.003) & 33854 & 0.9986 \\
\midrule
All categories & 0.2643 (0.0001) & 1.102 (0.001) & 925752 & 0.999 \\
\bottomrule
\end{tabular}}
\caption{Model by category, sorted by $\alpha$. Parentheses give clustered standard errors. Coefficients are close across all domains: $\alpha\in[0.262,0.266]$ ms/expert and $c\in[1.05,1.19]\times10^{-3}$ ms/token.}
\label{tab:appendix-model-categories}
\end{table*}

% compact appendix layout: barrier removed
\begin{figure*}[!tbp]
\centering
\includegraphics[width=0.86\textwidth]{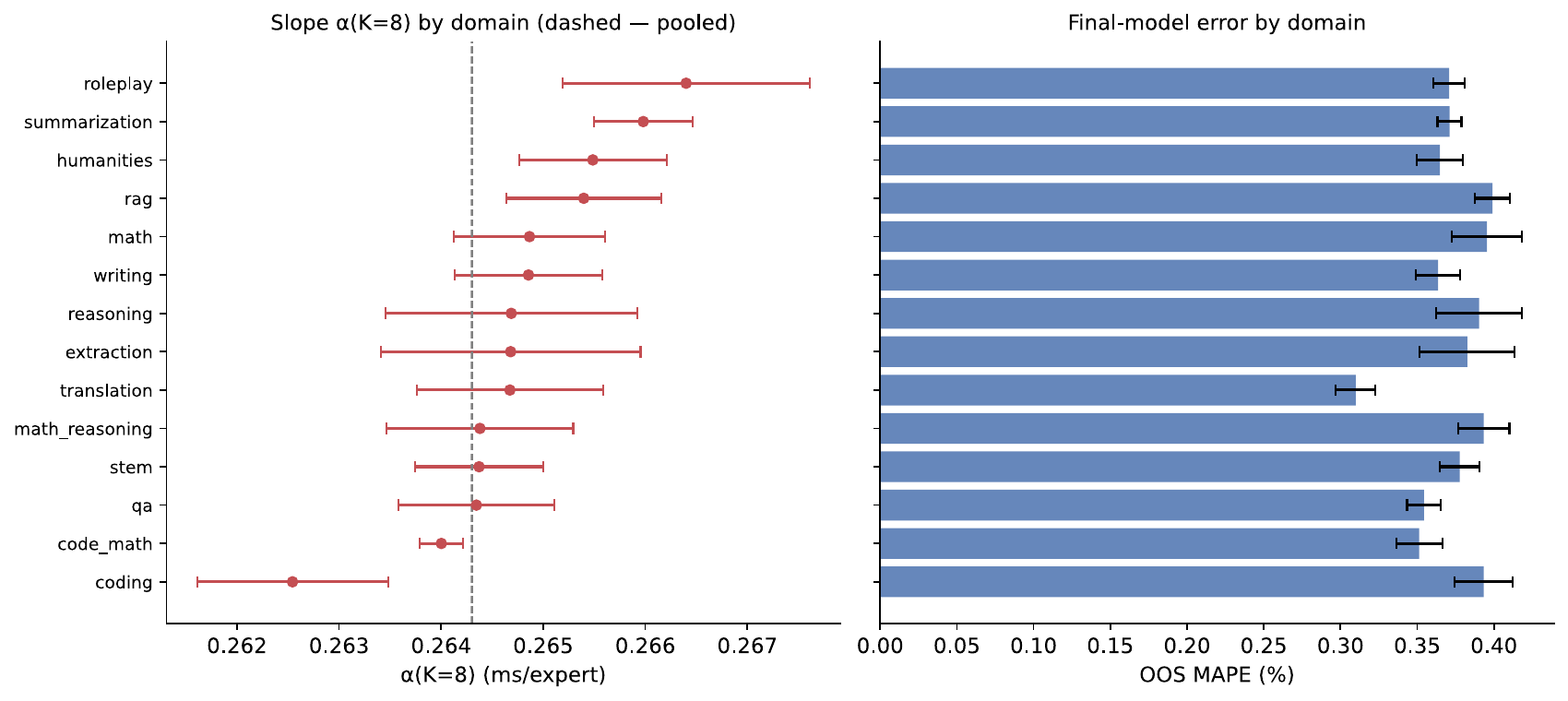}
\caption{Model coefficients by category: the slope $\alpha$ at $K=8$ with 95\% CI, and the error on held-out generations. The dashed line shows the estimate on the full pool.}
\label{fig:appendix-model-categories}
\end{figure*}

\begin{figure*}[!tbp]
\centering
\includegraphics[width=0.82\textwidth]{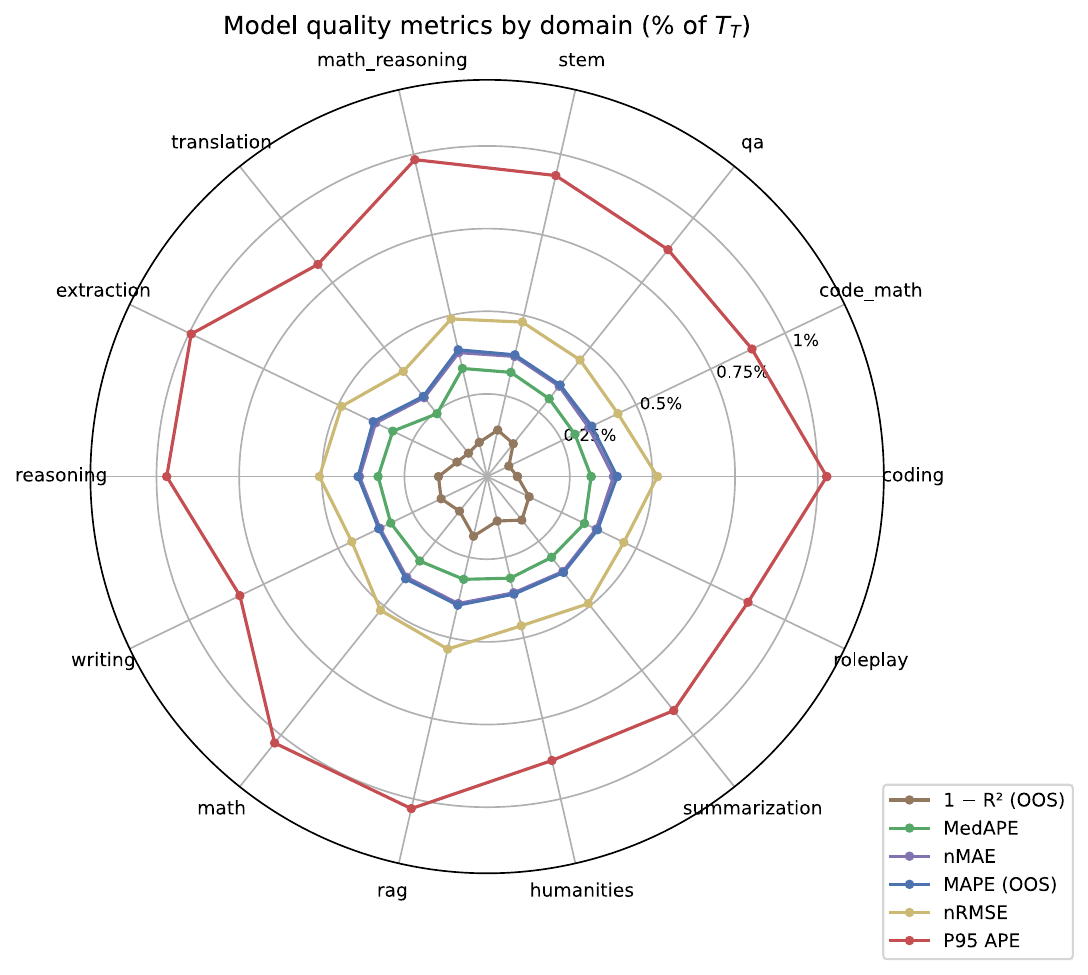}
\caption{Six dimensionless quality metrics by category: $1-R^2$, MedAPE, MAPE, nMAE, nRMSE, and P95 APE. Typical error is 0.3--0.4\%, and the 95th percentile does not exceed 1\%.}
\label{fig:appendix-model-error-radar}
\end{figure*}

% compact appendix layout: forced page break removed
\subsection{Budget \texorpdfstring{$K$}{K} as a Model Parameter}

The categorical specification was compared against a shared linear model in which the budget enters as an ordinary coefficient:
\begin{equation}
T_T
=
\beta_0
+
\alpha\bar U_r
+
cL
+
\gamma K
+
\varepsilon.
\label{eq:appendix-pooled-k-model}
\end{equation}
The model was fit on the full pool of $925\,752$ rounds; standard errors were clustered by generation, and out-of-sample metrics were computed via GroupKFold(5).
The estimated coefficients are $\alpha=0.2528\,(0.0014)$ ms/expert, $c=1.09\,(0.02)\times10^{-3}$ ms/token, $\gamma=-0.081\,(0.004)$ ms per unit of $K$, and $\beta_0=5.971\,(0.035)$ ms.
Under the clustered standard errors used, all four coefficients have $p<0.001$.
Because of the large number of nested rounds, the repetition of questions across $K$, and the choice of form after inspecting the sweep, these values describe a specific fit and are not used as causal or independent confirmation of the model.

\begin{table*}[!tbp]
\centering
\small
\begin{tabular}{@{}lrrr@{}}
\toprule
Model & $R^2$ & CV RMSE, ms & CV MAPE \\
\midrule
$\bar U_r+L$ & 0.856 & 0.685 & 4.42\% \\
$\bar U_r+L+\gamma K$ & 0.861 & 0.675 & 4.33\% \\
$\bar U_r\times C(K)+L$ & 0.999 & 0.056 & 0.37\% \\
\bottomrule
\end{tabular}
\caption{Budget as a linear coefficient versus the categorical form on the full pool; CV uses GroupKFold(5) by generation.}
\label{tab:appendix-budget-models}
\end{table*}

The linear coefficient on $K$ raises $R^2$ from 0.856 to 0.861 and reduces CV RMSE by $0.011\pm0.003$ ms, or 1.6\%.
The CV MAPE of the categorical form is roughly twelve times lower than that of the $\bar U_r+L$ model.

\begin{figure*}[!tbp]
\centering
\includegraphics[width=\textwidth]{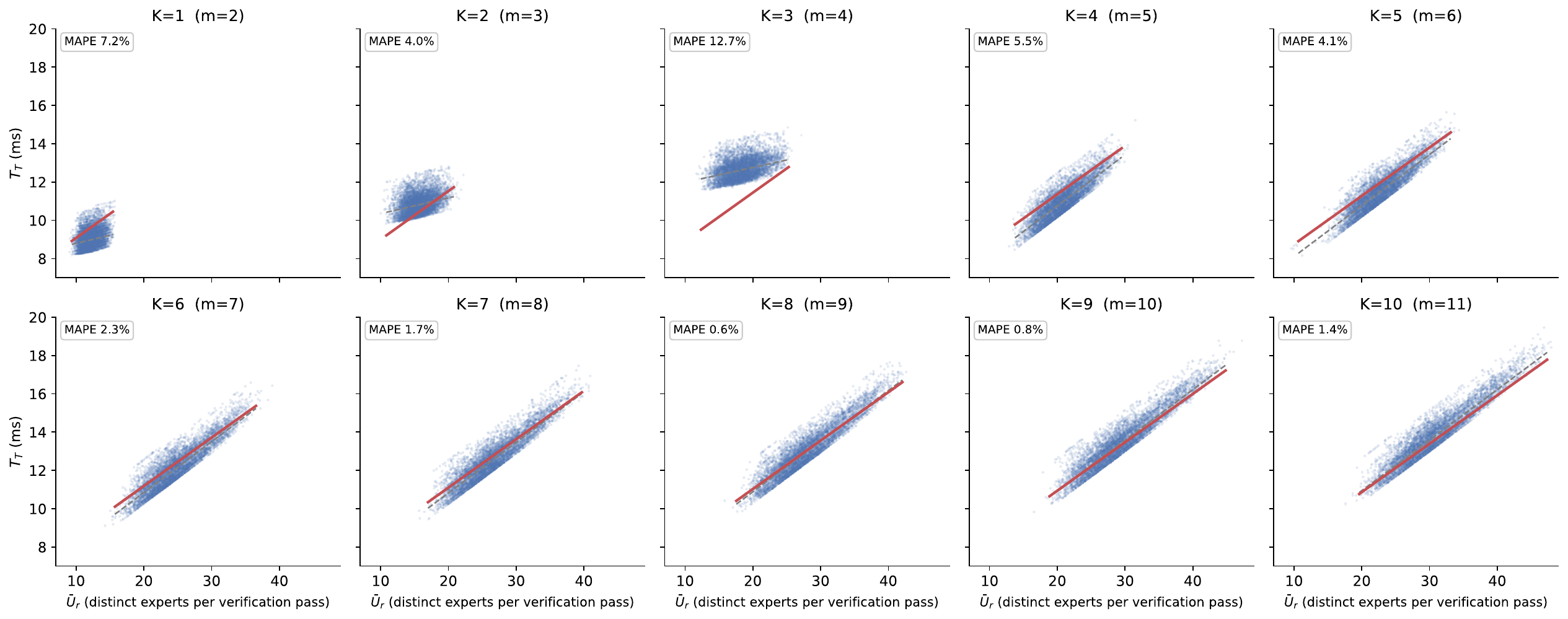}
\caption{The pooled model with a linear coefficient on $K$, at the median $L$ of each run. The red line shows the pooled model; the gray dashed lines show the separate per-$K$ fits. The pooled line reproduces the large-$m$ regime but is systematically off at $m\leq4$.}
\label{fig:appendix-k-model-scatter}
\end{figure*}

The pooled line tracks the large-regime clouds closely: MAPE is 0.6--4.1\% at $K\geq5$.
In the small regime, error reaches 7.2\% at $K=1$ and 12.7\% at $K=3$, since the pooled slope is dominated by the larger mass of large-regime rounds.

The standardized contribution of the budget is $\beta^*_K=-0.13$, versus $\beta^*_{\bar U}=0.99$ and $\beta^*_L=0.26$.
The negative $\gamma$ is a conditional coefficient arising from mixing regimes and is not interpreted as the marginal price of a position.

A leave-one-$K$-out evaluation of the pooled form gives 5.2\% MAPE on average, versus 0.26--0.59\% within a known $K$, with peaks of 12.7\% at $K=1$ and 15.1\% at $K=3$.
Other smooth parameterizations examined give 6.5--7.4\%, but their choice and breakpoint were not tuned within each training fold, so the comparison is exploratory rather than an independent validation of the categorical form.
Interpolating coefficients within the large regime achieves 0.26--0.65\% error, but transferring across the $m=4/5$ boundary gives 8.8--15.5\% error.
For the configuration studied, interpolation within the large regime requires fewer anchor values of $K$, whereas transfer across the observed boundary yields markedly larger error; this observation should not be transferred to another environment without recalibration.

% compact appendix layout: barrier removed
\subsection{Draft Time Model}

Draft time is modeled separately:
\begin{equation}
T_D=aK+b+\varepsilon_D.
\label{eq:appendix-draft-model}
\end{equation}
The OLS model was fit on all $925\,752$ rounds, with clustered standard errors and GroupKFold(5) by generation.
The estimates are $a=0.459\,(0.001)$ ms per step and $b=0.073\,(0.003)$ ms.
Under the clustered standard errors used, both coefficients have $p<0.001$; given the large $n$, the coefficient magnitudes and the held-out-generation error are more substantively important.
Mean $T_D$ by budget deviates from the line by 0.5\% on average, and the CV MedAPE of a typical round is 0.8\%.
Per-round $R^2=0.61$; the residual distribution has a heavy tail: 0.27\% of rounds have an absolute residual greater than 1 ms, and 0.3\% of records contain a zero timing.

\begin{figure*}[!tbp]
\centering
\includegraphics[width=0.82\textwidth]{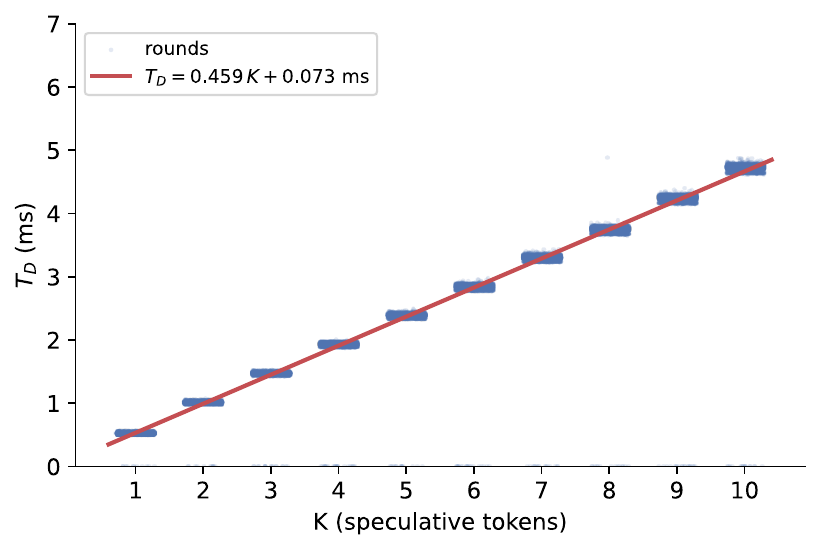}
\caption{Draft time versus budget and the model $T_D=0.459K+0.073$ ms. Each point corresponds to a round; points on the zero line represent records with a missing timing.}
\label{fig:appendix-draft-line}
\end{figure*}

The draft head is dense, so no routing term enters the model.
The mean dependence of $T_D$ on $K$ is approximately linear; the estimated slope is about 0.46 ms per step over the range $K=1,\dots,10$.
In this sweep, $T_D$ shows no visually apparent discreteness comparable to the change in $T_T$ at the $m=4/5$ boundary; a separate changepoint test for draft time was not performed.

% compact appendix layout: barrier removed
\subsection{Model Robustness}

A ladder of nested specifications shows that adding $L$ to $\bar U_r$ alone reduces CV MAPE from 5.6\% to 4.4\% and raises $R^2$ from 0.788 to 0.856.
Linear $m$, the interactions $mL$ and $m\bar U_r$, and $E_{\max}$ do not improve on a CV MAPE of 4.3\%.
Among the specifications examined, only the categorical budget form reduces CV MAPE to 0.37\% at $R^2=0.999$.

Estimation robustness was checked on the $K=8$ subsample in five ways: clustered bootstrap over 999 generation resamples, Huber regression, trimming 1\% of observations by absolute residual, separate fits on run halves, and leave-one-category-out across fourteen refits.
The estimate of $\alpha$ stays within 0.2642--0.2650 ms/expert, and $c$ within 1.100--1.105$\times10^{-3}$ ms/token.

\begin{table*}[!tbp]
\centering
\small
\resizebox{\textwidth}{!}{%
\begin{tabular}{@{}lrrr@{}}
\toprule
Check & $\alpha$ at $K=8$, ms/expert & $c$, $10^{-3}$ ms/token & $\beta_0$, ms \\
\midrule
OLS, clustered SE & 0.2643 [0.2642, 0.2645] & 1.1022 & 4.938 \\
Bootstrap by generation, 999 & 0.2643 [0.2642, 0.2645] & 1.1022 & 4.938 \\
Huber regression & 0.2644 & 1.1013 & 4.937 \\
1\% outlier trimming & 0.2644 & 1.1008 & 4.936 \\
First half of run & 0.2642 & 1.1003 & 4.942 \\
Second half of run & 0.2650 & 1.1046 & 4.919 \\
Leave-one-category-out & [0.2643, 0.2648] & [1.1009, 1.1037] & [4.926, 4.940] \\
\bottomrule
\end{tabular}}
\caption{Model coefficients under robustness checks on the $K=8$ subsample. For OLS and bootstrap, 95\% intervals are given; for leave-one-category-out, the range across fourteen refits.}
\label{tab:appendix-robustness}
\end{table*}

\begin{figure*}[!tbp]
\centering
\includegraphics[width=0.82\textwidth]{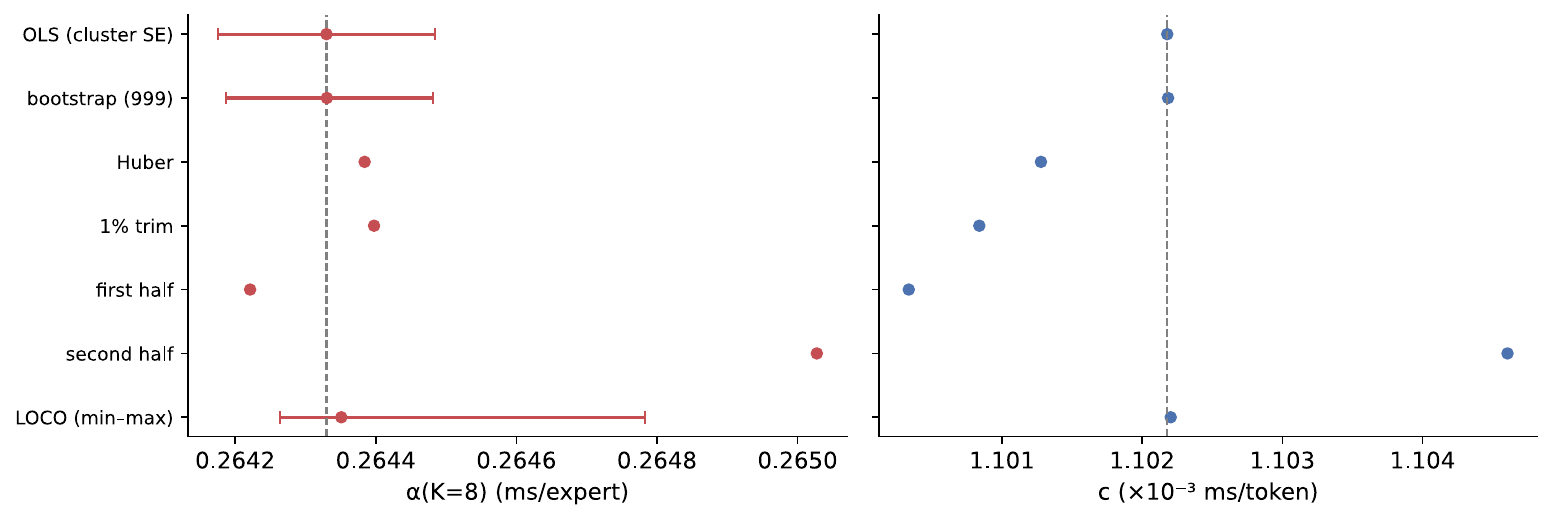}
\caption{Estimates of $\alpha$ at $K=8$ and $c$ under robustness checks. The dashed line marks the baseline OLS estimate; the entire spread stays within a fraction of a percent.}
\label{fig:appendix-robustness}
\end{figure*}

The context coefficient is close across the separate per-$K$ runs: $c\in[1.098,1.107]\times10^{-3}$, with variation below 0.9\%.
Mean residuals binned by $L$, KV-cache occupancy, and position within the generation show no trends.
Residual diagnostics indicate heteroscedasticity; the Breusch--Pagan test gives $p\approx10^{-28}$.
At $n\approx9\times10^5$, this value is treated as a diagnostic of residual shape rather than independent confirmation of a substantive effect; uncertainty is therefore assessed via clustered standard errors, with the emphasis on effect sizes, intervals, and OOS error.
At $n\approx9\times10^5$, normality of residuals is not required, and the low pairwise correlation $\operatorname{corr}(\bar U_r,L)=0.03$ does not indicate pronounced collinearity between these two covariates.

All coefficients and regime boundaries pertain only to the model, draft head, GPU, backend, CUDA-graph configuration, and $B=1$ studied.
A single seed and one full run per $K$ were used; run order, cross-session variance, and the causal status of the $m=4/5$ boundary were not tested with a randomized or reversed sweep.
The overhead of per-round tracing and double buffering was not measured separately.
Cross-hardware, cross-system, and batched transferability were not tested; the model requires recalibration when the environment changes or at $B>1$.
% compact appendix layout: barrier removed

%% file: appendix_linguistic_analysis_en.tex
\section{Linguistic Boundaries and Draft-Length Selection}
\label{app:linguistic-boundaries}

\newcommand{\Jfigure}[4][0.92]{%
  \begin{figure}[!tbp]
  \centering
  \includegraphics[width=#1\linewidth]{#2}
  \caption{#3}
  \label{#4}
  \end{figure}
}

\newcommand{\Jfloat}[4][0.88]{%
  \begin{figure}[!tbp]
  \centering
  \includegraphics[width=#1\linewidth]{#2}
  \caption{#3}
  \label{#4}
  \end{figure}
}

\newcommand{\Jwidefloat}[4][0.70]{%
  \begin{figure*}[!t]
  \centering
  \includegraphics[width=#1\textwidth]{#2}
  \caption{#3}
  \label{#4}
  \end{figure*}
}

\paragraph{Brief descriptive summary.}
The offline policy balances a surrogate match probability of sequential matching against an estimated computational cost, rather than simply maximizing the length of the matched prefix.
In this trace, regular fragments tend to receive a long draft, while a script change tends to receive a short one.
In four purposefully selected counterfactual cases, the surrogate match probability of the added match is 84.18--98.45\%, yet the full objective chooses a shorter boundary; after removing the expert term, the optimum shifts by one token in all four cases.
These cases were selected for exhibiting a shift and do not estimate the frequency of the effect.
Script change is also associated with the appearance of new experts in the code-domain-mixed subsample, whereas the first mismatch with the reference is not accompanied by a separate spike.

The premise is that a speculative draft is more useful where the next several tokens form a predictable continuation.
This appendix tests whether such regularity is reflected in the Oracle's decisions: whether it chooses a longer draft within stable fragments and shortens it near linguistic boundaries.

The analysis covers 39 responses and $9\,554$ decisions nested within them, at $L_{max}=8$.
Decisions within a single response are serially dependent; aggregated intervals are therefore computed with a clustered bootstrap at the response level.
At each position, the offline policy chooses a length $k^\star$ from one to eight tokens that minimizes the conditional expected cost.
This objective accounts for the probability of a reference-matching prefix and the computation of the draft and target models, including the activation of new experts.
$k^\star$ is the planned draft length, not the actual length of the matched prefix.
In the local examples, $a_{\mathrm{mode}}$ denotes the modal advancement length along the reconstructed trajectory.
The number of new experts shows how many experts, not previously used within a given continuation, the target MoE model brings in; the value is averaged across layers.

\subsection{What Is Tested}

Two related but distinct questions are tested.
\begin{enumerate}
    \item \textbf{Draft length.}
    If a continuation is linguistically regular --- for example, it repeats a code structure or completes an already-started construct --- the probability of matching several tokens with the reference should be higher.
    Length is chosen by the full Bellman equation, which includes the current cost and the continuation value; matching probability alone is not sufficient.
    Cases where the surrogate match probability and the cost point to different boundaries are therefore examined separately.
    \item \textbf{Expert routing.}
    If a transition between linguistic regimes requires a different set of computations, more new experts should appear at that position.
    A stronger version of the hypothesis predicts the same kind of spike at the first position of mismatch with the reference.
\end{enumerate}

For the first question, we compare code formatting and structure, ordinary text, sentence boundaries, transitions between code and text, and a switch from Latin to Cyrillic script.
We additionally examine a series of cases with a high probability of a matching prefix but a shorter cost-optimal length.
For the second question, we separately compare script change and the first draft mismatch.
This separation matters: a linguistic boundary, a mismatch with the reference, the surrogate match probability, and a change in the expert set are not the same event.

Each subsequent token is assigned to one local context: formatting, vocabulary, or code structure; vocabulary or a prose boundary; markup and formulas; a mode change; a script change.
For the selected examples, we additionally distinguish the start of a function, control flow, a comment, and a word continuation.
These labels describe the observed text fragment and are used only to compare groups.

\paragraph{Origin of the examples.}
Q1--Q6 correspond to $(qid,\,position,\,category)$: $(122,574,coding)$, $(66,408,code\_math)$, $(20,492,code\_math)$, $(7,38,code\_math)$, $(7,420,code\_math)$, and $(157,26,humanities)$.
A1--A12 correspond to $(qid,position)$: $(48,66)$, $(70,179)$, $(128,9)$, $(135,73)$, $(152,355)$, $(121,38)$, $(126,20)$, $(154,404)$, $(8,100)$, $(127,734)$, $(125,516)$, and $(52,332)$; their categories in order are: code\_math, code\_math, coding, extraction, humanities, coding, coding, humanities, code\_math, coding, coding, and code\_math.

\subsection{Local Examples}

Individual decisions show what phenomena hide behind the general ``boundary'' category.
We compare the continuations of the target and draft models, the chosen cut point $k^\star$, the first mismatch, and the modal advancement length $a_{\mathrm{mode}}$.
In the panels, the target model's continuation is shown on top and the draft model's below; matches, the first mismatch, and positions past the chosen boundary are distinguished by color, and lines mark $k^\star$ and $a_{\mathrm{mode}}$.

\subsubsection{Predictable Continuations and Major Boundaries}

\Jfigure{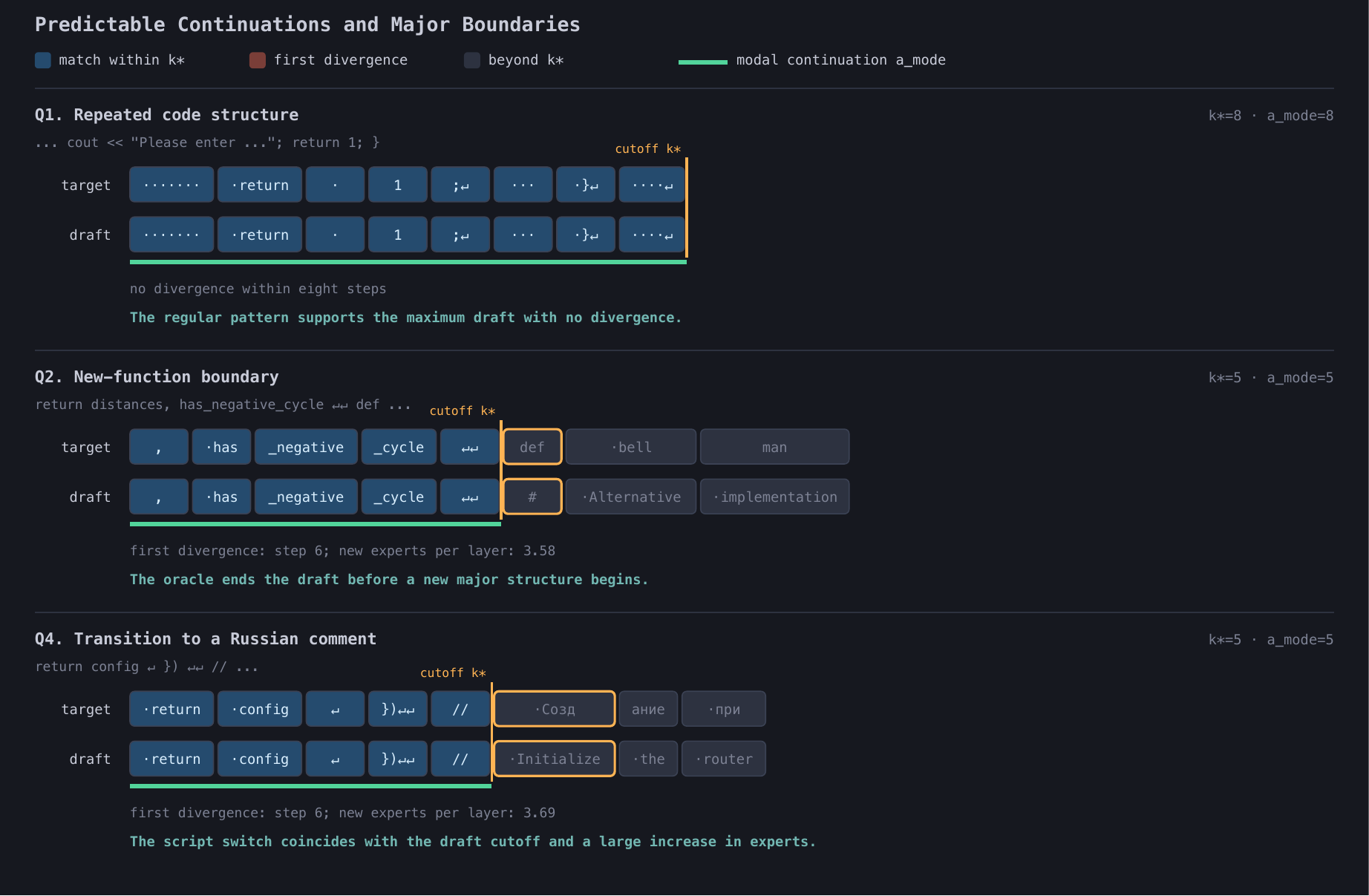}
{A regular fragment and two major boundaries. In the examples shown, a regular pattern coincides with a long $k^\star$, while a structural or language transition coincides with the stopping point.}
{fig:linguistic-local-boundaries}

In Q1, all eight tokens of a repeating pattern match; in this example the policy also chooses the maximum $k^\star=8$.
In Q2, the matching continuation ends right before the start of a new function, and in Q4 right before a transition from code to a Russian-language comment.
In both cases, the first mismatch lies past the chosen boundary.
In the examples shown, a regular pattern coincides with a long $k^\star$, while a structural or language transition coincides with the stopping point.

\subsubsection{Low-Cost Mismatches}

\Jfigure{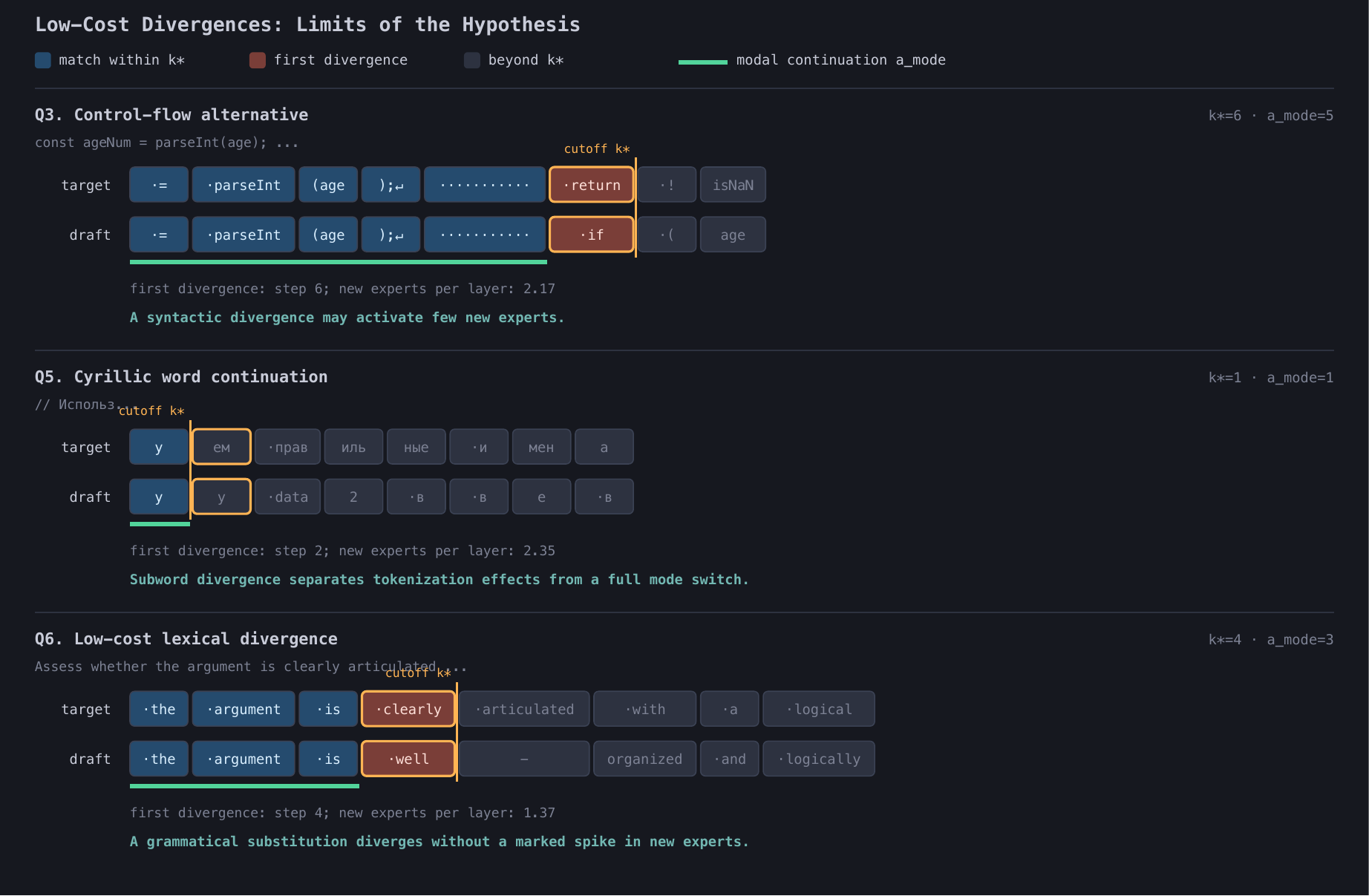}
{Three limitations of the simple boundary hypothesis. A token mismatch by itself does not imply a sharp change in routing, and the general label ``linguistic boundary'' does not determine the cost of a transition.}
{fig:linguistic-local-limits}

In Q3, the draft proposes \texttt{if} instead of \texttt{return}; both continuations are syntactically valid, and the number of new experts stays small.
In Q5, the mismatch occurs inside a Cyrillic word and is partly determined by tokenization.
In Q6, replacing \texttt{clearly} with \texttt{well} is grammatical and is accompanied by only 1.37 new experts per layer.
Hence, a token mismatch by itself does not imply a sharp change in routing, and the general label ``linguistic boundary'' does not determine the cost of a transition.

\subsection{Does the Pattern Generalize Across the Full Trace?}

The local examples explain individual decisions but do not show how often each effect occurs.
A comparison across all decisions tests the first hypothesis: whether the chosen length $k^\star$ differs across linguistic contexts.

\Jfigure{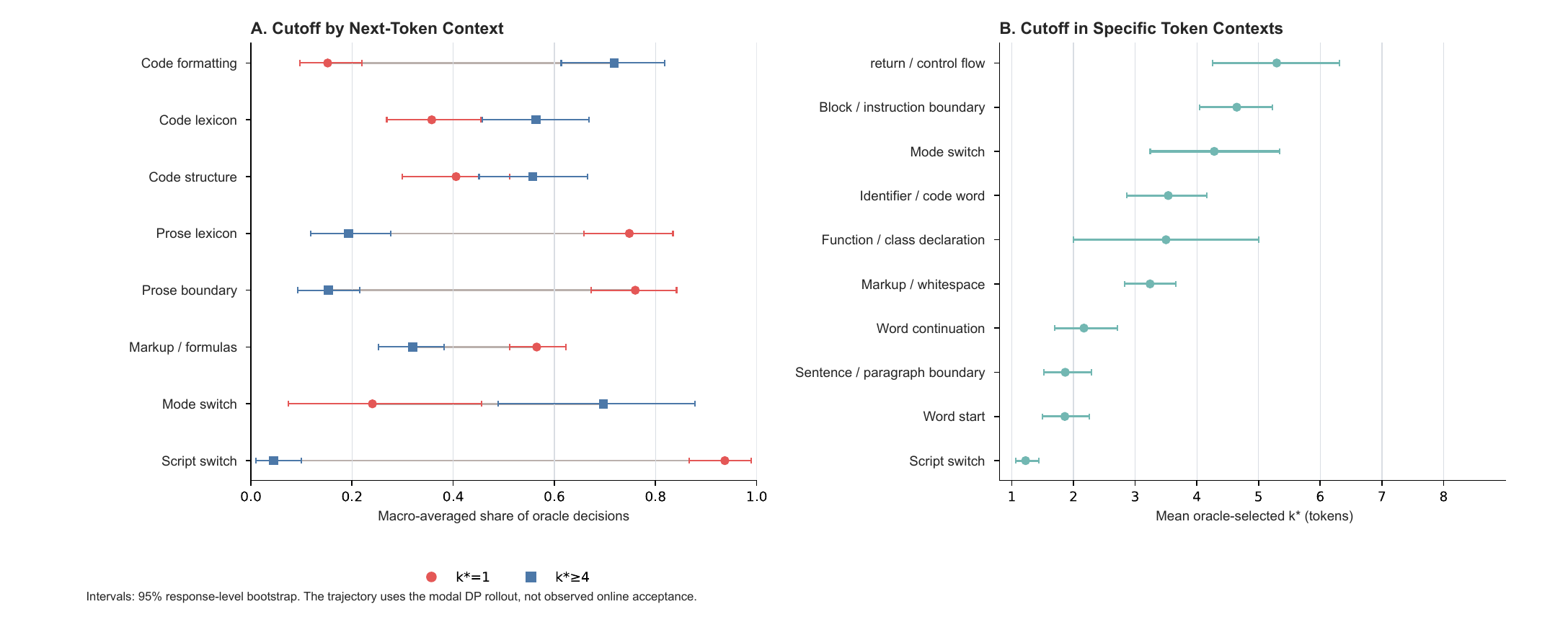}
{Context-wise distributions of the chosen length. On the left, the shares of $k^\star=1$ and $k^\star\geq4$; on the right, the mean $k^\star$ by boundary type. Intervals are a 95\% clustered bootstrap at the response level and reflect between-response variability.}
{fig:linguistic-contexts}

The strongest contrast is observed between code formatting and script change.
For code formatting, the share of $k^\star\geq4$ is 0.718 $[0.614;0.818]$ ($455$ decisions, $22$ responses); for prose vocabulary it is 0.193 $[0.119;0.276]$ ($4383/30$); and for script change it is 0.045 $[0.010;0.100]$ ($68/8$).
At script change, the share of $k^\star=1$ is 0.937 $[0.867;0.990]$.

However, not every boundary leads to a short draft.
At transitions between code and ordinary text, the descriptive share of decisions with $k^\star\geq4$ is 0.697.
Return and control-flow keywords are also associated with a longer, rather than shorter, continuation.

In the studied sample, we therefore observe a descriptive association between regular formatting and a long draft, and between script change and a short one.
But the general category ``linguistic boundary'' is not sufficient by itself: the specific type of transition matters.

\subsection{When High Match Probability Does Not Imply a Long Draft}

The aggregated analysis shows where the policy chooses a long draft, but it does not separate two possible reasons for stopping: a drop in match probability and a rise in computational cost.
Below we present four cases in which the adjacent longer sequence remains linguistically coherent and has a high surrogate match probability.

For each case, we first consider the decision of the full objective function.
Then the coefficient on expert cost is zeroed out, $\alpha=0$, and the entire dynamic-programming problem is solved again.
The remaining terms are kept, and the future cost is recomputed for the new policy.
The counterfactual ablation recomputes the policy value in the four selected cases and thereby tests the contribution of the expert term at the level of the Oracle's objective function.
It is not a causal test of the hardware mechanism and does not estimate the frequency of the effect.

\Jfigure{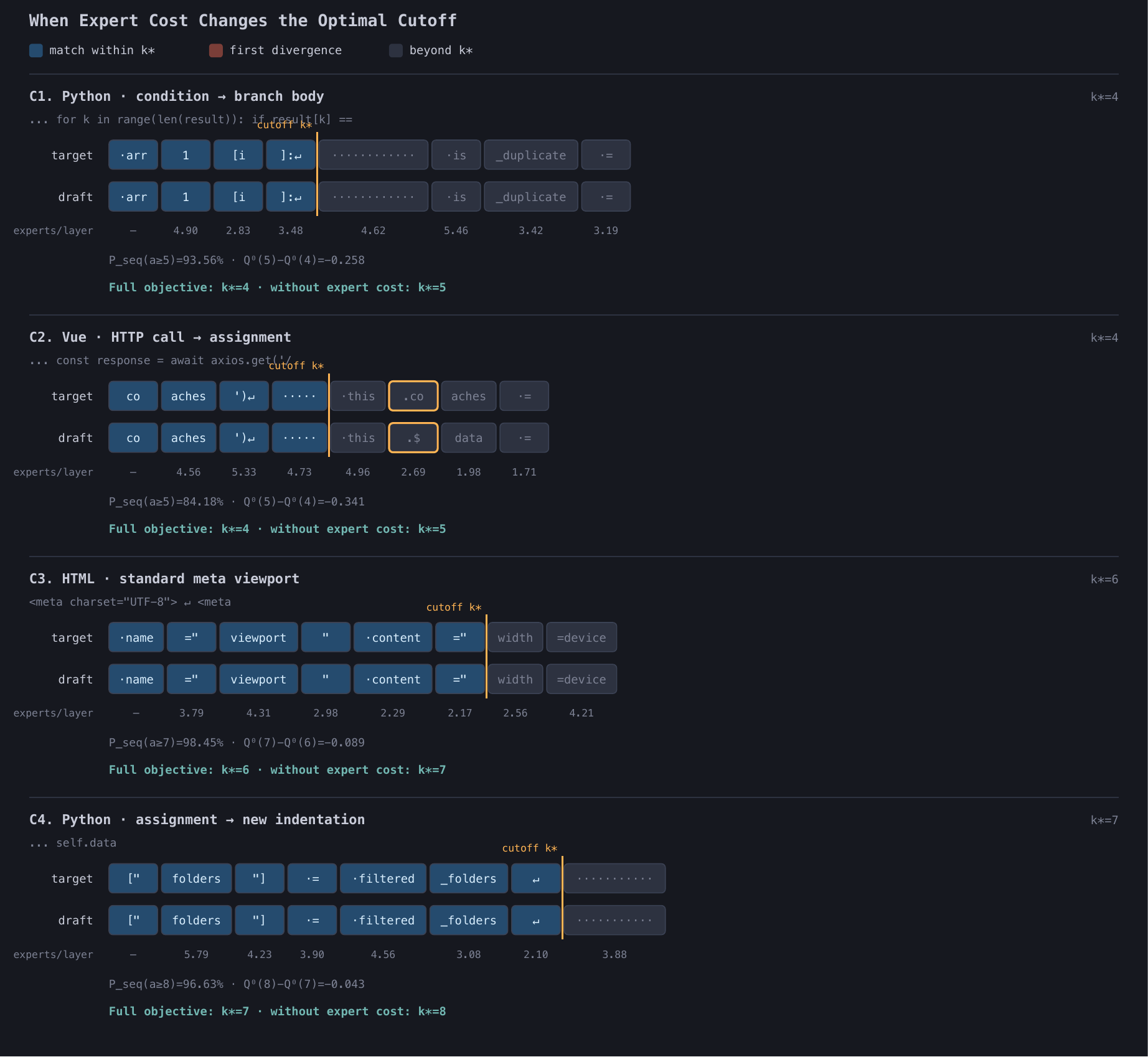}
{Four cases where expert cost shifts the boundary. In all cases, the full objective chooses $k^\star$, while without expert cost the neighboring length $k^\star+1$ becomes cheaper. The local increase in new experts serves as a diagnostic, not a standalone causal test.}
{fig:linguistic-cost-cutoff}

C1 completes a Python condition and moves into the branch body; C2 completes an HTTP call and begins an assignment.
C3 passes through a \texttt{meta viewport} construct, and C4 completes a list and moves to a new indented line.
These descriptions capture the linguistic context of four purposefully selected examples but do not turn them into a frequency-representative sample.

The four cases were selected for exhibiting a shift after ablation and span the boundaries $4\rightarrow5$, $6\rightarrow7$, and $7\rightarrow8$.
The surrogate match probability of the added token ranges from 84.18\% to 98.45\%, and all four differences $Q^0(k^\star+1)-Q^0(k^\star)$ are negative.
Hence, the sum of the remaining terms favors a longer draft, and the expert term shifts the resulting boundary.

The saved trace does not contain the full cost matrix, the value function, or all the $Q(k)$ terms, so it cannot be used to precisely decompose the full difference between neighboring lengths or to attribute the decision to a single local jump in the expert count.
Only a local conclusion follows from them: without the expert term, the policy chooses a different length in all four illustrations; the prevalence of the effect across the full trace is not estimated.

\subsection{Are Boundaries Associated with New Experts?}

We next test the second hypothesis.
If a complex linguistic boundary changes the MoE model's routing, more new experts should appear at that position.
If the same mechanism is directly linked to mismatches, a similar spike should be observed at the first position of mismatch with the reference.

\Jfigure{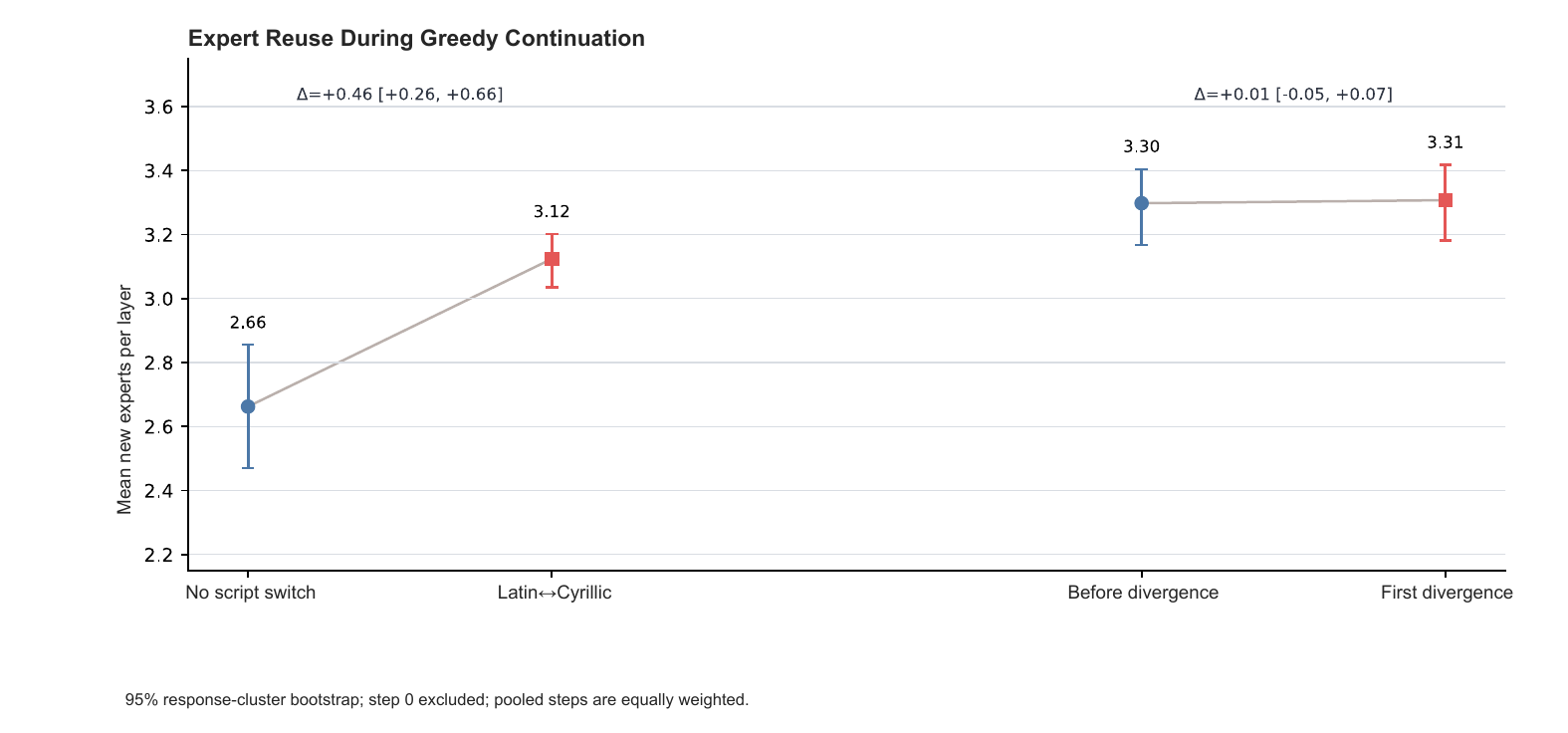}
{New experts at script change and at the first draft mismatch. The left pair compares positions without a script change to Latin$\leftrightarrow$Cyrillic transitions; the right pair compares the position before a mismatch with the first mismatch position. A higher value means the continuation brings in more previously unused experts. Intervals are a 95\% clustered bootstrap at the response level.}
{fig:linguistic-expert-reuse}

At a Latin $\leftrightarrow$ Cyrillic transition, the number of new experts increases from 2.66 to 3.12 experts per layer.
The difference is $+0.46$, and the 95\% interval $[+0.26;+0.66]$ excludes zero.
This difference is consistent with an association between script change and routing but does not separate script from the code domain.

At the first mismatch position, the picture is different: 3.31 experts versus 3.30 at the preceding step.
The difference is $+0.01$, and the interval $[-0.05;+0.07]$ includes zero.
No separate spike in new experts is observed at the first mismatch position.

Comparisons are aligned by continuation step, and uncertainty is estimated with a clustered bootstrap at the response level.
The result is compatible with the weak associative hypothesis linking script change to routing but does not support the causal version in which the appearance of new experts directly explains mismatches.

\subsection{Interpretation and Limitations}

The descriptive analysis yields four observations.
\begin{enumerate}
    \item Regular formatting and repeated code structure are associated with a longer draft.
    \item A high probability of a matching prefix does not guarantee a long draft: expert cost can shift the conditional boundary.
    \item In the code-domain-mixed subsample, script change is associated with a short $k^\star$ and the appearance of new experts.
    \item The first mismatch position by itself is not accompanied by an observable change in the number of experts.
\end{enumerate}

These observations point to a relationship between linguistic structure, draft-length choice, and expert routing, but they do not establish a causal mechanism.
The analysis was conducted for a single target/draft model pair and 39 responses.
Cyrillic examples are concentrated mainly in code-containing responses, so the effect of script cannot be fully separated from the domain.
Distances between hidden representations were not measured.

Linguistic structure is linked to two different signals.
In the selected examples, regularity is accompanied by a high surrogate match probability; the aggregated effect size is not estimated.
$k^\star$ is determined by the full Bellman equation, which includes transition probabilities, current cost, and the continuation value; therefore, neither linguistic coherence nor a full draft match by itself determines the chosen length.
Taken together, the selected examples show the policy's sensitivity to the type and predictability of the continuation, rather than a simple linguistic rule.
A repeated structure can go together with a long draft; a major boundary with a stop; a local mismatch with a small increase in experts with continuation; a predictable script change with either a short or a maximum $k^\star$.

\subsection{Additional Qualitative Examples}

This gallery extends the local analysis and shows how different the decisions with the same $k^\star$ can be.
The sample is stratified, not random.
Three groups were defined first: regular continuations, structural boundaries, and counterexamples to the simple rule ``a boundary or a mismatch with the reference implies early truncation and many new experts.''
The first group includes exact eight-token matches of different types; the second includes linguistically interpretable transitions; the third includes cases that directly limit the simple hypothesis.
All 12 examples are drawn from different responses and do not repeat the six local examples above.
The gallery is used for interpretation, not to estimate the frequency or magnitude of the effect.

For each example, we compare the continuations of the target and draft models, the cut point, and the number of new experts at that step.
The value of the first step is not used for the linguistic contrast, since it equals eight by construction.

\Jfloat{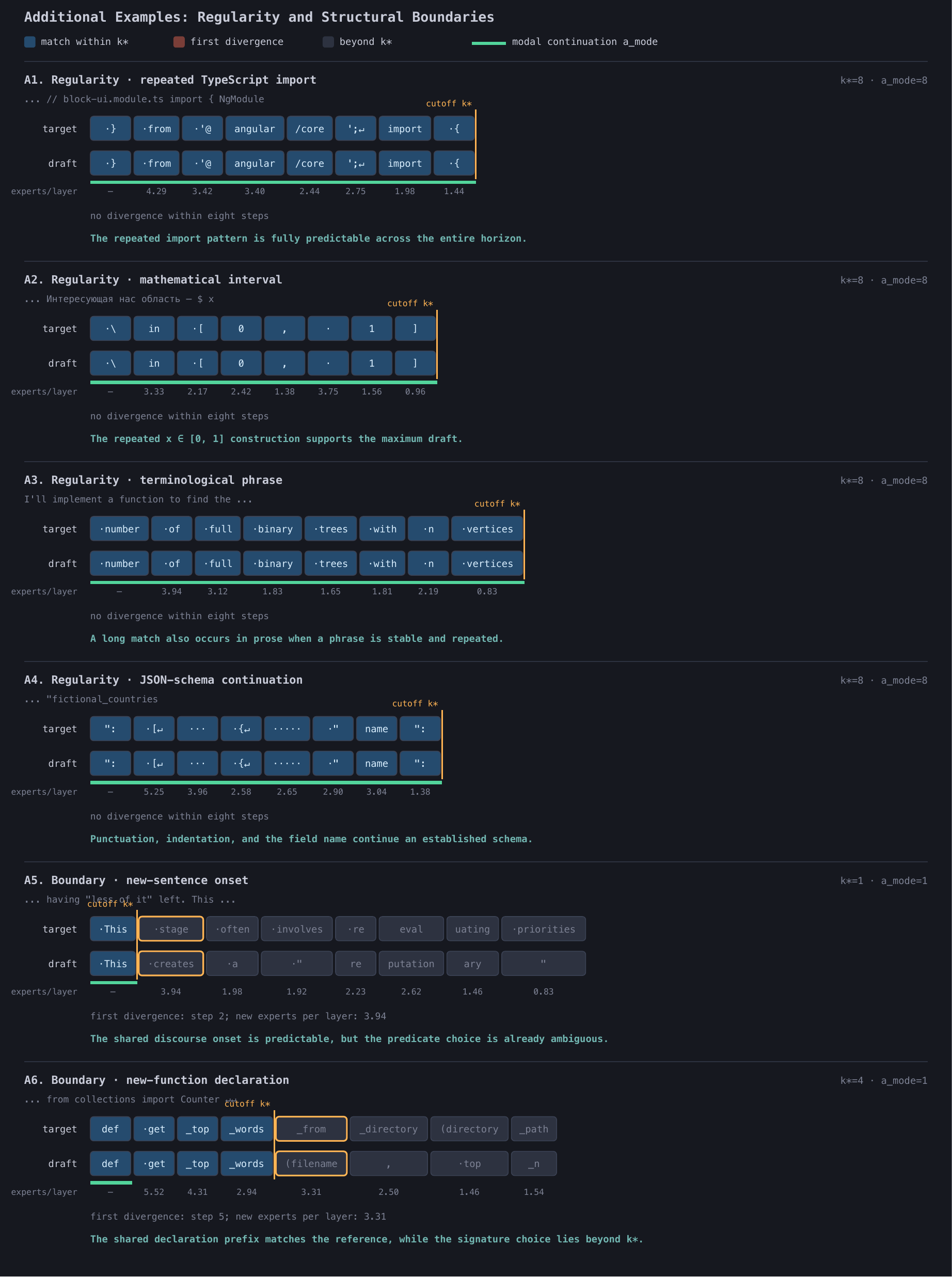}
{Four selected eight-token matches from different genres and two structural boundaries.}
{fig:linguistic-additional-regular}

\Jfloat{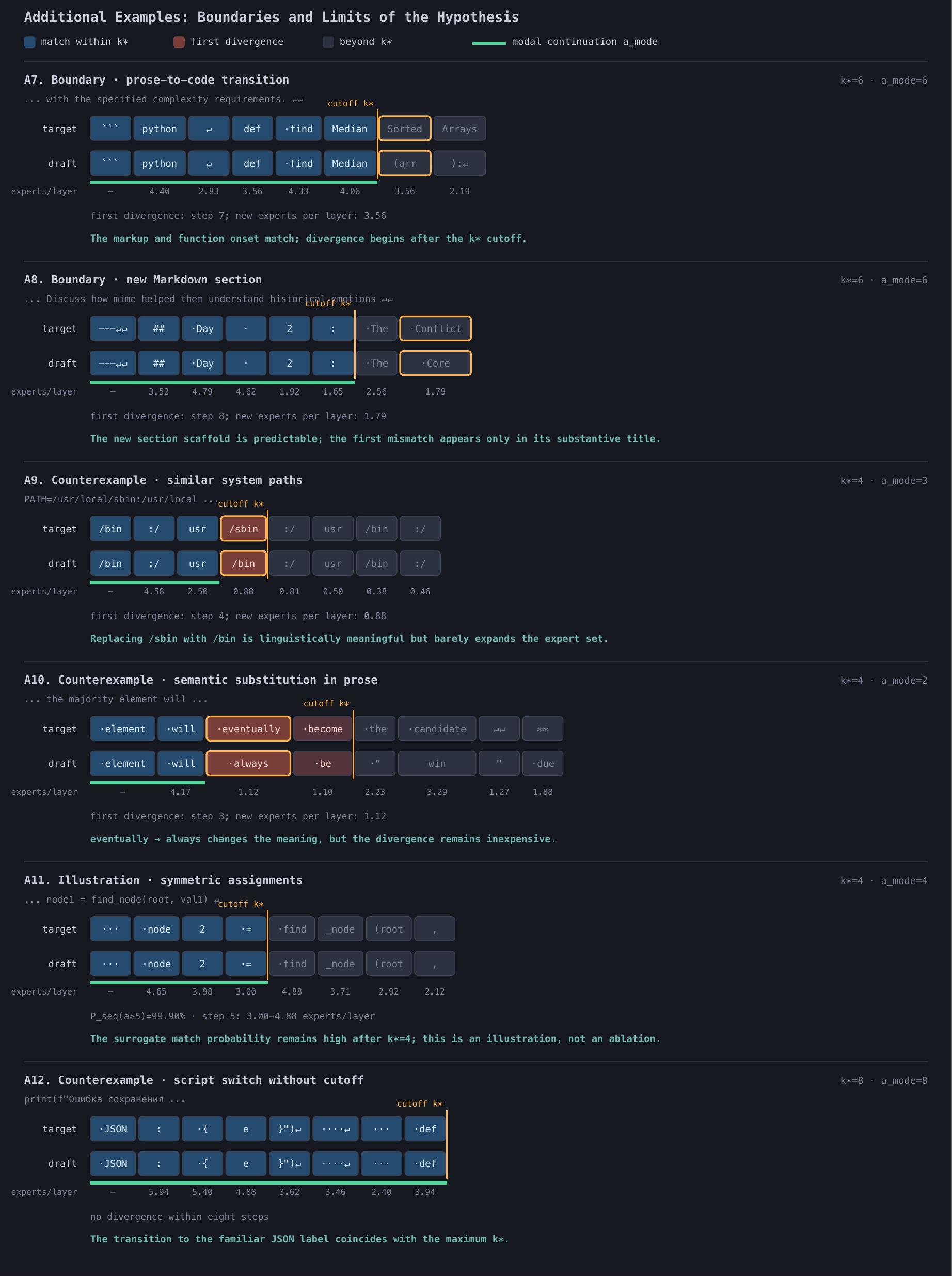}
{Two major boundaries and four counterexamples. Neither a formal boundary, nor a full match, nor a semantic mismatch alone determines the Oracle's decision.}
{fig:linguistic-additional-limits}

\subsubsection{Regular Continuations and Initial Structural Boundaries}

A1--A4 show a full eight-token match in a TypeScript import, a mathematical interval, a stable prose phrase, and a JSON schema.
These examples show that full eight-token matches occur across several genres; their frequency is not estimated.
In A5, the Oracle retains only the general discourse opener \texttt{This} and truncates the draft before the choice of predicate.
In A6, the header of a new function matches, while the choice of its signature begins already past $k^\star$.

\subsubsection{Boundaries and Limitations of the Simple Hypothesis}

In A7, a transition from prose to code, and in A8, a new Markdown section, both have a predictable structural scaffold; the substantive mismatch arises after the chosen boundary.
In A9, replacing \texttt{/sbin} with \texttt{/bin}, and in A10, \texttt{eventually} with \texttt{always}, occur within the draft but bring in few new experts.

In A11, the surrogate match probability of matching at least five tokens is 99.90\%, yet the policy chooses $k^\star=4$; the local rise in the number of new experts makes the cost-based interpretation plausible, but the $\alpha=0$ ablation does not change the decision here.
In A12, a switch from Cyrillic to the familiar technical token \texttt{JSON} coincides with the maximum $k^\star=8$.
Hence, neither a formal boundary, nor a full match, nor a semantic mismatch alone determines the Oracle's decision.

Additional examples illustrate the diversity of decisions but do not estimate the prevalence of the patterns.
They include repeated structures, major boundaries, low-cost mismatches, and predictable script changes.

%% file: main_arxiv.bbl
\begin{thebibliography}{31}
\providecommand{\natexlab}[1]{#1}

\bibitem[{Abramovich et~al.(2026)Abramovich, Ashkenazi, Putterman, Chislett,
  Mitra, Darvish~Rouhani, Zilberstein, and Geifman}]{abramovich2026speedbench}
Talor Abramovich, Maor Ashkenazi, Izzy Putterman, Benjamin Chislett, Tiyasa
  Mitra, Bita Darvish~Rouhani, Ran Zilberstein, and Yonatan Geifman. 2026.
\newblock \href {https://doi.org/10.48550/arXiv.2604.09557} {{SPEED-Bench}: A
  unified and diverse benchmark for speculative decoding}.
\newblock \emph{Preprint}, arXiv:2604.09557.

\bibitem[{Bang et~al.(2026)Bang, Cho, Hwang, Chung, and Rhu}]{bang2026specmoe}
Jehyeon Bang, Eunyeong Cho, Ranggi Hwang, Jinha Chung, and Minsoo Rhu. 2026.
\newblock \href {https://arxiv.org/abs/2604.10152} {{SpecMoE}: A fast and
  efficient mixture-of-experts inference via self-assisted speculative
  decoding}.
\newblock \emph{Preprint}, arXiv:2604.10152.
\newblock Extended version accepted at DAC 2026.

\bibitem[{Cai et~al.(2024)Cai, Li, Geng, Peng, Lee, Chen, and
  Dao}]{cai2024medusa}
Tianle Cai, Yuhong Li, Zhengyang Geng, Hongwu Peng, Jason~D. Lee, Deming Chen,
  and Tri Dao. 2024.
\newblock \href {https://proceedings.mlr.press/v235/cai24b.html} {Medusa:
  Simple {LLM} inference acceleration framework with multiple decoding heads}.
\newblock In \emph{Proceedings of the 41st International Conference on Machine
  Learning}, volume 235 of \emph{Proceedings of Machine Learning Research},
  pages 5209--5235. PMLR.

\bibitem[{Chen et~al.(2023)Chen, Borgeaud, Irving, Lespiau, Sifre, and
  Jumper}]{chen2023acceleratinglargelanguagemodel}
Charlie Chen, Sebastian Borgeaud, Geoffrey Irving, Jean-Baptiste Lespiau,
  Laurent Sifre, and John Jumper. 2023.
\newblock \href {https://arxiv.org/abs/2302.01318} {Accelerating large language
  model decoding with speculative sampling}.
\newblock \emph{Preprint}, arXiv:2302.01318.

\bibitem[{Chen et~al.(2025)Chen, Wen, Wu, Zhang, and Wu}]{chen2025spmoe}
Liangkun Chen, Zijian Wen, Tian Wu, Xiaoxi Zhang, and Chuan Wu. 2025.
\newblock \href {https://arxiv.org/abs/2510.10302} {{SP-MoE}: Speculative
  decoding and prefetching for accelerating {MoE}-based model inference}.
\newblock \emph{Preprint}, arXiv:2510.10302.

\bibitem[{Fedus et~al.(2022)Fedus, Zoph, and Shazeer}]{fedus2022switch}
William Fedus, Barret Zoph, and Noam Shazeer. 2022.
\newblock \href {https://arxiv.org/abs/2101.03961} {Switch transformers:
  Scaling to trillion parameter models with simple and efficient sparsity}.
\newblock \emph{Journal of Machine Learning Research}, 23(120):1--39.

\bibitem[{Gautam et~al.(2025)Gautam, Shrestha, and Reddy}]{gautam2025gammatune}
Aayush Gautam, Susav Shrestha, and Narasimha Reddy. 2025.
\newblock \href {https://arxiv.org/abs/2504.00030} {Token-driven {GammaTune}:
  Adaptive calibration for enhanced speculative decoding}.
\newblock \emph{Preprint}, arXiv:2504.00030.

\bibitem[{Huang et~al.(2025)Huang, Zhu, Zhan, Hu, Mao, Yu, Liu, and
  Zhang}]{huang2025moesd}
Zongle Huang, Lei Zhu, Zongyuan Zhan, Ting Hu, Weikai Mao, Xianzhi Yu, Yongpan
  Liu, and Tianyu Zhang. 2025.
\newblock \href {https://doi.org/10.48550/arXiv.2505.19645} {{MoESD}: Unveil
  speculative decoding's potential for accelerating sparse {MoE}}.
\newblock \emph{Preprint}, arXiv:2505.19645.
\newblock Accepted as spotlight at NeurIPS 2025.

\bibitem[{Lepikhin et~al.(2020)Lepikhin, Lee, Xu, Chen, Firat, Huang, Krikun,
  Shazeer, and Chen}]{lepikhin2020gshard}
Dmitry Lepikhin, HyoukJoong Lee, Yuanzhong Xu, Dehao Chen, Orhan Firat, Yanping
  Huang, Maxim Krikun, Noam Shazeer, and Zhifeng Chen. 2020.
\newblock \href {https://arxiv.org/abs/2006.16668} {{GShard}: Scaling giant
  models with conditional computation and automatic sharding}.
\newblock \emph{arXiv preprint arXiv:2006.16668}.

\bibitem[{Leviathan et~al.(2023)Leviathan, Kalman, and
  Matias}]{pmlr-v202-leviathan23a}
Yaniv Leviathan, Matan Kalman, and Yossi Matias. 2023.
\newblock \href {https://proceedings.mlr.press/v202/leviathan23a.html} {Fast
  inference from transformers via speculative decoding}.
\newblock In \emph{Proceedings of the 40th International Conference on Machine
  Learning}, volume 202 of \emph{Proceedings of Machine Learning Research},
  pages 19274--19286. PMLR.

\bibitem[{Li et~al.(2024{\natexlab{a}})Li, Wei, Zhang, and
  Zhang}]{li2024eagle2}
Yuhui Li, Fangyun Wei, Chao Zhang, and Hongyang Zhang. 2024{\natexlab{a}}.
\newblock \href {https://doi.org/10.48550/arXiv.2406.16858} {{EAGLE-2}: Faster
  inference of language models with dynamic draft trees}.
\newblock \emph{Preprint}, arXiv:2406.16858.

\bibitem[{Li et~al.(2024{\natexlab{b}})Li, Wei, Zhang, and Zhang}]{li2024eagle}
Yuhui Li, Fangyun Wei, Chao Zhang, and Hongyang Zhang. 2024{\natexlab{b}}.
\newblock \href {https://proceedings.mlr.press/v235/li24bt.html} {{EAGLE}:
  Speculative sampling requires rethinking feature uncertainty}.
\newblock In \emph{Proceedings of the 41st International Conference on Machine
  Learning}, volume 235 of \emph{Proceedings of Machine Learning Research},
  pages 28935--28948. PMLR.

\bibitem[{Li et~al.(2025)Li, Wei, Zhang, and Zhang}]{li2025eagle3}
Yuhui Li, Fangyun Wei, Chao Zhang, and Hongyang Zhang. 2025.
\newblock \href {https://doi.org/10.48550/arXiv.2503.01840} {{EAGLE-3}: Scaling
  up inference acceleration of large language models via training-time test}.
\newblock \emph{Preprint}, arXiv:2503.01840.

\bibitem[{Mamou et~al.(2024)Mamou, Pereg, Korat, Berchansky, Timor, Wasserblat,
  and Schwartz}]{pmlr-v262-mamou24a}
Jonathan Mamou, Oren Pereg, Daniel Korat, Moshe Berchansky, Nadav Timor, Moshe
  Wasserblat, and Roy Schwartz. 2024.
\newblock \href {https://proceedings.mlr.press/v262/mamou24a.html} {Dynamic
  speculation lookahead accelerates speculative decoding of large language
  models}.
\newblock In \emph{Proceedings of The 4th NeurIPS Efficient Natural Language
  and Speech Processing Workshop}, volume 262 of \emph{Proceedings of Machine
  Learning Research}, pages 456--467. PMLR.

\bibitem[{McDanel et~al.(2026)McDanel, Li, Surineni, and
  Khaitan}]{mcdanel2026moespec}
Bradley McDanel, Steven Li, Sruthikesh Surineni, and Harshit Khaitan. 2026.
\newblock \href {https://arxiv.org/abs/2602.16052} {{MoE-Spec}: Expert
  budgeting for efficient speculative decoding}.
\newblock \emph{Preprint}, arXiv:2602.16052.

\bibitem[{Pan et~al.(2026)Pan, Tao, Pang, Wang, Zhao, and Zhang}]{pan2026evict}
Lehan Pan, Ziyang Tao, Ruoyu Pang, Xiao Wang, Jianjun Zhao, and Yanyong Zhang.
  2026.
\newblock \href {https://arxiv.org/abs/2605.00342} {Making every verified token
  count: Adaptive verification for {MoE} speculative decoding}.
\newblock \emph{Preprint}, arXiv:2605.00342.

\bibitem[{Pope et~al.(2023)Pope, Douglas, Chowdhery, Devlin, Bradbury, Heek,
  Xiao, Agrawal, and Dean}]{MLSYS2023_c4be71ab}
Reiner Pope, Sholto Douglas, Aakanksha Chowdhery, Jacob Devlin, James Bradbury,
  Jonathan Heek, Kefan Xiao, Shivani Agrawal, and Jeff Dean. 2023.
\newblock \href
  {https://proceedings.mlsys.org/paper_files/paper/2023/file/c4be71ab8d24cdfb45e3d06dbfca2780-Paper-mlsys2023.pdf}
  {Efficiently scaling transformer inference}.
\newblock In \emph{Proceedings of Machine Learning and Systems}, volume~5,
  pages 606--624.

\bibitem[{Samarin et~al.(2026)Samarin, Krutikov, Shevtsov, Skvortsov, Fisin,
  and Golubev}]{samarin2026lklosses}
Alexander Samarin, Sergei Krutikov, Anton Shevtsov, Sergei Skvortsov, Filipp
  Fisin, and Alexander Golubev. 2026.
\newblock \href {https://doi.org/10.48550/arXiv.2602.23881} {{LK} losses:
  Direct acceptance rate optimization for speculative decoding}.
\newblock \emph{Preprint}, arXiv:2602.23881.

\bibitem[{Saxena et~al.(2025)Saxena, Tsai, Taneja, Jaleel, and
  Qureshi}]{saxena2025utility}
Anish Saxena, Po-An Tsai, Hritvik Taneja, Aamer Jaleel, and Moinuddin Qureshi.
  2025.
\newblock \href {https://doi.org/10.48550/arXiv.2506.20675} {Utility-driven
  speculative decoding for mixture-of-experts}.
\newblock \emph{Preprint}, arXiv:2506.20675.

\bibitem[{Shazeer(2019)}]{shazeer2019fast}
Noam Shazeer. 2019.
\newblock \href {https://arxiv.org/abs/1911.02150} {Fast transformer decoding:
  One write-head is all you need}.
\newblock \emph{arXiv preprint arXiv:1911.02150}.

\bibitem[{Shazeer et~al.(2017)Shazeer, Mirhoseini, Maziarz, Davis, Le, Hinton,
  and Dean}]{shazeer2017outrageously}
Noam Shazeer, Azalia Mirhoseini, Krzysztof Maziarz, Andy Davis, Quoc Le,
  Geoffrey Hinton, and Jeff Dean. 2017.
\newblock \href {https://arxiv.org/abs/1701.06538} {Outrageously large neural
  networks: The sparsely-gated mixture-of-experts layer}.
\newblock In \emph{International Conference on Learning Representations}.

\bibitem[{Vaswani et~al.(2017)Vaswani, Shazeer, Parmar, Uszkoreit, Jones,
  Gomez, Kaiser, and Polosukhin}]{NIPS2017_3f5ee243}
Ashish Vaswani, Noam Shazeer, Niki Parmar, Jakob Uszkoreit, Llion Jones,
  Aidan~N. Gomez, Lukasz Kaiser, and Illia Polosukhin. 2017.
\newblock \href
  {https://proceedings.neurips.cc/paper_files/paper/2017/file/3f5ee243547dee91fbd053c1c4a845aa-Paper.pdf}
  {Attention is all you need}.
\newblock In \emph{Advances in Neural Information Processing Systems},
  volume~30. Curran Associates, Inc.

\bibitem[{Wang et~al.(2025{\natexlab{a}})Wang, Tan, Hu, Qin, Sun, Xie, Cai, Li,
  and Zhang}]{wang2025sparseverification}
Jikai Wang, Jianchao Tan, Yuxuan Hu, Jiayu Qin, Yerui Sun, Yuchen Xie, Xunliang
  Cai, Juntao Li, and Min Zhang. 2025{\natexlab{a}}.
\newblock \href {https://arxiv.org/abs/2512.21911} {Accelerate speculative
  decoding with sparse computation in verification}.
\newblock \emph{Preprint}, arXiv:2512.21911.

\bibitem[{Wang et~al.(2025{\natexlab{b}})Wang, Liu, Hou, Xia, Tang, Zhang, Li,
  and Guo}]{wang2025moespeq}
Wenfeng Wang, Jiacheng Liu, Xiaofeng Hou, Xinfeng Xia, Peng Tang, Mingxuan
  Zhang, Chao Li, and Minyi Guo. 2025{\natexlab{b}}.
\newblock \href {https://arxiv.org/abs/2511.14102} {{MoE-SpeQ}: Speculative
  quantized decoding with proactive expert prefetching and offloading for
  mixture-of-experts}.
\newblock \emph{Preprint}, arXiv:2511.14102.

\bibitem[{Wang et~al.(2026)Wang, Chen, Zhen, Liu, Zheng, Liu, Xu, and
  Li}]{wang2026prism}
Xuliang Wang, Yuetao Chen, Maochan Zhen, Fang Liu, Xinzhou Zheng, Xingwu Liu,
  Hong Xu, and Ming Li. 2026.
\newblock \href
  {https://proceedings.mlsys.org/paper_files/paper/2026/hash/414fd191b3246a19a55741b938380136-Abstract-Conference.html}
  {{PRISM}: Parametrically refactor inference for speculative decoding draft
  models}.
\newblock In \emph{Proceedings of Machine Learning and Systems}, volume~8.

\bibitem[{Wang et~al.(2025{\natexlab{c}})Wang, Zhang, Zhou, Wang, Zhou, Jiang,
  Cai, Huan, Gu, Zhong, and Tian}]{wang2025specmoeoff}
Zhibin Wang, Zhonghui Zhang, Yuhang Zhou, Zibo Wang, Mo~Zhou, Peng Jiang,
  Weilin Cai, Chengying Huan, Rong Gu, Sheng Zhong, and Chen Tian.
  2025{\natexlab{c}}.
\newblock \href {https://arxiv.org/abs/2508.21706} {Accelerating
  mixture-of-experts inference by hiding offloading latency with speculative
  decoding}.
\newblock \emph{Preprint}, arXiv:2508.21706.

\bibitem[{Wu et~al.(2025)Wu, Zhou, Verma, Prakash, Rus, and Low}]{wu2025tetris}
Zhaoxuan Wu, Zijian Zhou, Arun Verma, Alok Prakash, Daniela Rus, and Bryan
  Kian~Hsiang Low. 2025.
\newblock \href {https://doi.org/10.48550/arXiv.2502.15197} {{TETRIS}: Optimal
  draft token selection for batch speculative decoding}.
\newblock \emph{Preprint}, arXiv:2502.15197.

\bibitem[{Xie et~al.(2026)Xie, Liu, Huang, Ling, Dai, Zheng, and
  Hu}]{xie2026lessexperts}
Jincheng Xie, Runheng Liu, Heyan Huang, Yawen Ling, Hanbin Dai, Yu~Zheng, and
  Wen Hu. 2026.
\newblock \href {https://doi.org/10.48550/arXiv.2607.12696} {Less experts,
  faster decoding: Cost-aware speculative decoding for mixture-of-experts}.
\newblock \emph{Preprint}, arXiv:2607.12696.

\bibitem[{Zhang et~al.(2024{\natexlab{a}})Zhang, Wang, Ma, Zhu, Chen, Lan, and
  Yu}]{zhang2024adaeagle}
Situo Zhang, Hankun Wang, Da~Ma, Zichen Zhu, Lu~Chen, Kunyao Lan, and Kai Yu.
  2024{\natexlab{a}}.
\newblock \href {https://doi.org/10.48550/arXiv.2412.18910} {{AdaEAGLE}:
  Optimizing speculative decoding via explicit modeling of adaptive draft
  structures}.
\newblock \emph{Preprint}, arXiv:2412.18910.

\bibitem[{Zhang et~al.(2024{\natexlab{b}})Zhang, Xu, Liang, Chen, He, Wang, and
  Tu}]{zhang2024draftmodelknows}
Ziyin Zhang, Jiahao Xu, Tian Liang, Xingyu Chen, Zhiwei He, Rui Wang, and
  Zhaopeng Tu. 2024{\natexlab{b}}.
\newblock \href {https://doi.org/10.48550/arXiv.2411.18462} {Draft model knows
  when to stop: Self-verification speculative decoding for long-form
  generation}.
\newblock \emph{Preprint}, arXiv:2411.18462.

\bibitem[{Zhong et~al.(2025)Zhong, Bharadwaj, Wang, Ji, and
  Lee}]{zhong2025beagle}
Wei Zhong, Manasa Bharadwaj, Yixiao Wang, Yipeng Ji, and Chul Lee. 2025.
\newblock \href {https://doi.org/10.48550/arXiv.2505.24544} {Cross-attention
  speculative decoding}.
\newblock \emph{Preprint}, arXiv:2505.24544.

\end{thebibliography}
